\documentclass[compsoc,conference,a4paper,10pt,times]{IEEEtran}

\usepackage[utf8]{inputenc}
\usepackage{amsmath}
\usepackage{amssymb}
\usepackage{amsthm}
\usepackage{xcolor}
\usepackage{mathtools}
\usepackage{soul}
\usepackage{url}  
\usepackage{algorithm}
\usepackage{tcolorbox}
\usepackage[noend]{algpseudocode}
\usepackage{subcaption}
\usepackage{hyperref}
\usepackage{enumitem}
\usepackage{booktabs} 
\usepackage{color}
\usepackage{multirow} 
\usepackage{siunitx} 
\usepackage{amsfonts} 
\newcommand{\NA}{{--}} 
\usepackage[numbers]{natbib}
\usepackage{multirow}
\newcommand\blfootnote[1]{%
  \begingroup
  \renewcommand\thefootnote{}\footnote{#1}%
  \addtocounter{footnote}{-1}%
  \endgroup
}

\usepackage{todonotes}
\usepackage{booktabs}
\usepackage{soul}

\usepackage{todonotes}
\usepackage{comment}

\newif\ifdraft
\drafttrue
\usepackage{xparse}
\usepackage{xspace}

\makeatletter
\def\@addpunct2#1{\ifnum\spacefactor>\@m \else#1\fi}
\newcommand{\para}[1]{\noindent\textbf{#1\unskip\@addpunct2{.}}~~}
\makeatother

\newcommand{\dataset}{$D$\xspace}
\newcommand{\model}{$M$\xspace}

\newcommand{\weights}{\textbf{w}\xspace}
\newcommand{\fingerprint}[1][]{}
\NewDocumentCommand{\pw}{o}{
  \IfNoValueTF{#1}
    {\mathcal{P}}
    {\mathcal{P}(#1)}%
}
\newcommand{\prf}{PoL\xspace}
\MakeRobust\prf

\NewDocumentCommand{\oracle}{o}{
  \IfNoValueTF{#1}
    {Or}
    {Or(#1)}%
}
\NewDocumentCommand{\verifier}{o}{
  \IfNoValueTF{#1}
    {\mathcal{V}}
    {\mathcal{V}(#1)}%
}

\newcommand{\redst}[1][115pt]{\bgroup\markoverwith {\textcolor{red}{\makebox[0pt][l]{\rule[0.5ex]{#1}{0.4pt}}\rule[1ex]{#1}{0.4pt}}}\ULon}

\newcommand{\ie}{\textit{i.e.,}\@\xspace}
\newcommand{\eg}{\textit{e.g.,}\@\xspace}
\newcommand{\etal}{\textit{et al.}\@\xspace}

\algnewcommand\algorithmicoptional{\textbf{Optional:}}
\algnewcommand\Optional{\item[\algorithmicoptional]}%

\ifdraft
\newcommand{\nick}[1]{\textcolor{blue}{nick: #1}}
\newcommand{\varun}[1]{\textcolor{cyan}{VC: #1}}
\newcommand{\chris}[1]{\textcolor{teal}{CC: #1}}

\newcommand{\ilia}[1]{\textcolor{green}{ilia: #1}}
\newcommand{\anvinth}[1]{\textcolor{orange}{anvith: #1}}
\newcommand{\my}[1]{\textcolor{purple}{M.: #1}}
\newcommand{\natalie}[1]{{\color{red} [natalie: #1]}}

\newcommand{\adam}[1]{{\color{orange} [adam: #1]}}
\newcommand{\vinith}[1]{{\color{brown} [vinith: #1]}}
\newcommand{\lucy}[1]{{\color{lightgray} [lucy: #1]}}
\else
\newcommand{\nick}[1]{}
\newcommand{\varun}[1]{}
\newcommand{\chris}[1]{}
\newcommand{\ilia}[1]{}
\newcommand{\anvinth}[1]{}
\newcommand{\my}[1]{}
\newcommand{\natalie}[1]{}
\newcommand{\stephan}[1]{}
\newcommand{\adam}[1]{}
\newcommand{\vinith}[1]{}
\newcommand{\lucy}[1]{}
\fi

\newtheorem{lemma}{Lemma}
\newtheorem{corollary}{Corollary}

\title{Unrolling SGD: Understanding Factors Influencing Machine Unlearning}

\author{Anvith Thudi$^{\dag}$\text{*}, Gabriel Deza$^{\dag}$\text{*}, Varun Chandrasekaran, Nicolas Papernot$^{\dag}$ \\ University of Toronto and Vector Institute $\dag$, University of Wisconsin-Madison}

\definecolor{lightgray}{gray}{0.75}

\begin{document}

\maketitle

\begin{abstract}

Machine unlearning is the process through which a deployed machine learning model \textcolor{black}{is made to forget} about \textcolor{black}{some} of its training data points. While naively retraining the model from scratch is an option, it is almost always associated with large computational overheads for deep learning models. Thus, several approaches to {\em approximately unlearn} have been proposed along with corresponding metrics that formalize what it means for a model to forget about a data point. In this work, we first taxonomize approaches and metrics of approximate unlearning. As a result, we identify {\em verification error}, \ie the $\ell_2$ difference between the weights of an approximately unlearned and a naively retrained model, as an approximate unlearning metric that should be optimized for as it subsumes a large class of other metrics. We theoretically analyze the canonical training algorithm, stochastic gradient descent (SGD), to surface the variables which are relevant to reducing the verification error of approximate unlearning for SGD. From this analysis, we first derive an easy-to-compute proxy for verification error (termed {\em unlearning error}). The analysis also informs the design of a new training objective penalty that limits the overall change in weights during SGD and as a result facilitates approximate unlearning with lower verification error. We validate our theoretical work through an empirical evaluation on learning with CIFAR-10,  CIFAR-100, and IMDB sentiment analysis.

\end{abstract}

\blfootnote{\text{*}Equal Contribution}

\section{Introduction}
\label{sec:intro}

The goal of machine unlearning is to provide a mechanism for removing the impact a datapoint in the training set had on the final model. This is motivated by multiple settings where forgetting a datapoint is paramount. For example, a model which has not unlearned may leak some of the private information contained in a point~\cite{shokri2017membership, abadi2016deep}. This is particularly relevant in scenarios where a particular user who owns the data later revokes access to it--a possibility popularized by the {\em right-to-be-forgotten} in the GDPR~\cite{mantelero2013eu}. The \textcolor{black}{odds} of leakage is exacerbated by the ability of a ML model to memorize parts of its dataset~\cite{carlini2019secret,fredrikson2015model}.

Machine unlearning was first introduced by Cao \etal~\cite{cao2015towards} to unlearn datapoints for simple hypothesis spaces which have known SQ learning algorithms. Machine unlearning has since been extended to deep neural networks (DNNs)~\cite{bourtoule2019machine, graves2020amnesiac,golatkar2020eternal,sekhari2021remember,guo2019certified,baumhauer2020machine,scrubbingv2}, \textcolor{black}{and its privacy implications have been studied~\cite{10.1145/3460120.3484756,gupta2021adaptive,10.1145/3372297.3417880}.} There are currently two broad approaches to machine unlearning: retraining and approximate unlearning. 

With retraining, the point to be unlearned is removed from the training set, and a new model is trained from scratch on this updated training set. This approach has the advantage of carrying a strong claim for why the new model was not influenced in any way by the point to be unlearned, as it is not trained on it. This is a non-trivial advantage when unlearning needs to be transparent, \eg in the context of right-to-be-forgotten requests~\cite{mantelero2013eu}. Naively retraining from scratch can be made more efficient~\cite{bourtoule2019machine}. Nevertheless it is still an expensive process that requires changes to the training pipeline. 

In approximate unlearning, the model owner instead starts from the existing model and seeks to modify its weights so as to obtain a slightly different model which satisfies an unlearning \textit{criterion} (\eg poor performance on the point to be unlearned, or {\em similar} weights to a model naively retrained without the point to be unlearned, etc). This can be done for example through a form of gradient ascent~\cite{graves2020amnesiac}, which is the opposite of gradient descent performed to train the model normally. Alternatively various approaches suggest hessian-based updates~\cite{guo2019certified,golatkar2020eternal}. Approximate unlearning has the advantage of being more computationally efficient than retraining but comes at the expense of a weaker guarantee: the model learnt may not be completely un-influenced by the unlearned datapoint.

Underlying the ambiguity with the various approximate claims to unlearning is simply the variety of such unlearning criterion that individual approaches consider and their associated metrics: each typically has their own metric for measuring unlearning, but it is not clear how to compare claims made with different approaches/metrics. Ideally, there would be one metric which captures most, if not all the intuitive properties of unlearning covered by metrics proposed to this day. Our work tackles this problem by first showing that \textit{verification error}---the $\ell_2$ difference in weights between an approximately unlearned model and a naively retrained model---implies a large class of the other metrics. Thus it carries a stronger intuition, and helps unify the pursuits of other work. 

Nevertheless, verification error has its own faults. To measure it, one needs to compute the naively retrained model, which begs the question of why not just use the retrained model as the unlearned model? Furthermore, despite discounting randomness associated with training algorithms, training a model is noisy due to numerical instabilities introduced by back-end randomness in floating point computations; comparison with respect to any one retrained model leaves the metric itself noisy. It is also not immediately apparent what one should do then to reduce the verification error. Additionally, for verification error to subsume other metrics, the unlearning method should satisfy specific properties (refer \S~\ref{sec: ver_err_implies_weights}).

Our work answers these questions. First by \textcolor{black}{expanding} the canonical training algorithm, \textcolor{black}{stochastic gradient descent (SGD)}, with a Taylor series and then further analyzing it, we formalize an unlearning method---{\em single gradient unlearning}---that only depends on the initial weights (and obtain some other terms that capture approximation error); unlearning is {\em approximate} (due to the error term) but {\em inexpensive} as it depends on only the initial weights.

Second, the analysis leads to an approximation of an unlearned model's verification error which we call the \textit{unlearning error}. The main advantage of this approximation compared to verification error is that it does not require computing the retrained model (\ie the ground truth that one would obtain by retraining from scratch to unlearn a point). This makes unlearning error cheaper to compute, and also alleviates the issue of the retrained model being noisy in the computation of verification error. We empirically show that unlearning error is strongly correlated with verification error in relevant contexts. This allows us to envision directly minimizing unlearning error during training, so as to improve our ability to later unlearn with smaller verification error. This is particularly appealing given that unlearning error is expressed in a form where the variables we need to change during training to decrease unlearning error are apparent. 

These properties of unlearning error help us (and future efforts building on our work) design approximate unlearning mechanisms with lower verification error. Specifically, we propose our {\em standard deviation (SD) loss} which we show effectively decreases this unlearning error (and \textcolor{black}{consequently,} verification error). Intuitively, SD loss forces the model to converge with less overall change to the weights. This allows us to unlearn a point from models trained with the SD loss using single gradient unlearning.

To summarize, the main contributions of our work are:
\begin{enumerate}
\item A taxonomy of approximate unlearning which concludes with verification error as a metric to study as it subsumes a large class of unlearning criteria. 
\item An analysis of SGD which (a) introduces an inexpensive mechanism for unlearning termed single gradient unlearning, and (b) uncovers the variables impacting verification error. This not only yields an easier-to-compute proxy for verification error, but also informs how to train models that are easier to unlearn well. 
\item A way of decreasing this unlearning error (and thus in turn verification error) by the use of our SD loss with little impact to performance. We validate our approach empirically \textcolor{black}{for models trained on} on CIFAR-10, CIFAR-100, and IMDb sentiment classification. 
\end{enumerate}

\section{Primer on Deep Learning \& Notation}
\label{sec:intro_dl}



In our work, we focus on supervised learning~\cite{hastie2009overview}. We utilize a bold-faced font to denote vectors (and tensors). We consider a dataset \dataset which consists of pairs $\{(\mathbf{x}_i,y_i)_{i \in [n]}\}$ (where $[n] = \{0,1,\cdots,n-1\}$); $\mathbf{x}$ is a datapoint (\eg an image) and $y$ is its label. We wish to train a model $M$, which is a parameterized function of weights \weights that we can modify (or learn); mathematically, the model is denoted as the function $M: \mathcal{X} \rightarrow \mathcal{Y}$, where $\mathcal{X}$ denotes the space of inputs (\ie $\mathbf{x} \in \mathcal{X}$) and $\mathcal{Y}$ denotes the space of outputs (\ie $y \in \mathcal{Y}$). Ideally, the model should be denoted $M_{\mathbf{w}}$, but we omit the dependence on $\mathbf{w}$ when the context is clear. Our experiments consider deep neural networks (DNNs) given their success on various difficult tasks~\cite{krizhevsky2012imagenet}, and their large training costs. 


To learn the weights that make $M$ best classify \dataset, we minimize a loss function $\mathcal{L}$ that measures the error our model has when predicting the label $y$ from an input $\mathbf{x}$. Examples of such loss functions include the cross-entropy (CE) loss~\cite{cox1958regression} which is the de-facto choice for classification tasks. The cross-entropy loss is defined as:
\begin{equation}
    \mathcal{L}_{CE}(M(\mathbf{x}),\mathbf{y}) = - \sum_{i=1}^{c} y_{i}\log(M(\mathbf{x})_{i})
\end{equation}
where $M(\mathbf{x})$ returns a probability simplex, and $M(\mathbf{x})_i$ is the $i^{th}$ value in this simplex, and $\mathbf{y} = (y_1, \cdots, y_c)$ s.t. $y_i \in \{0,1\}$ is the one-hot vector defining the label of $\mathbf{x}$, and $c$ is the number of classes. Since a closed form solution cannot be found analytically for non-convex models such as DNNs, the weights $\mathbf{w}$ are learnt in an iterative manner. Many established optimization schemes are derived from mini-batch stochastic gradient descent (SGD)~\cite{kiefer1952stochastic}. Formally speaking, mini-batch SGD can be described as below:
\begin{equation*}
\weights_{t+1} = \weights_t - \eta \frac{\partial \mathcal{L}}{\partial \weights}|_{\weights_t,\hat{\mathbf{x}}_t}
\end{equation*}
where weights at step $t$ are obtained using the weights from step $t-1$, $\hat{\mathbf{x}}_t$ is the mini-batch of data used at step $t$, and $\eta$ denotes the learning rate hyperparameter.

\section{Taxonomy of Approximate Unlearning}
\label{sec:overview_unlearning}

\begin{figure}
    \includegraphics[width = \columnwidth]{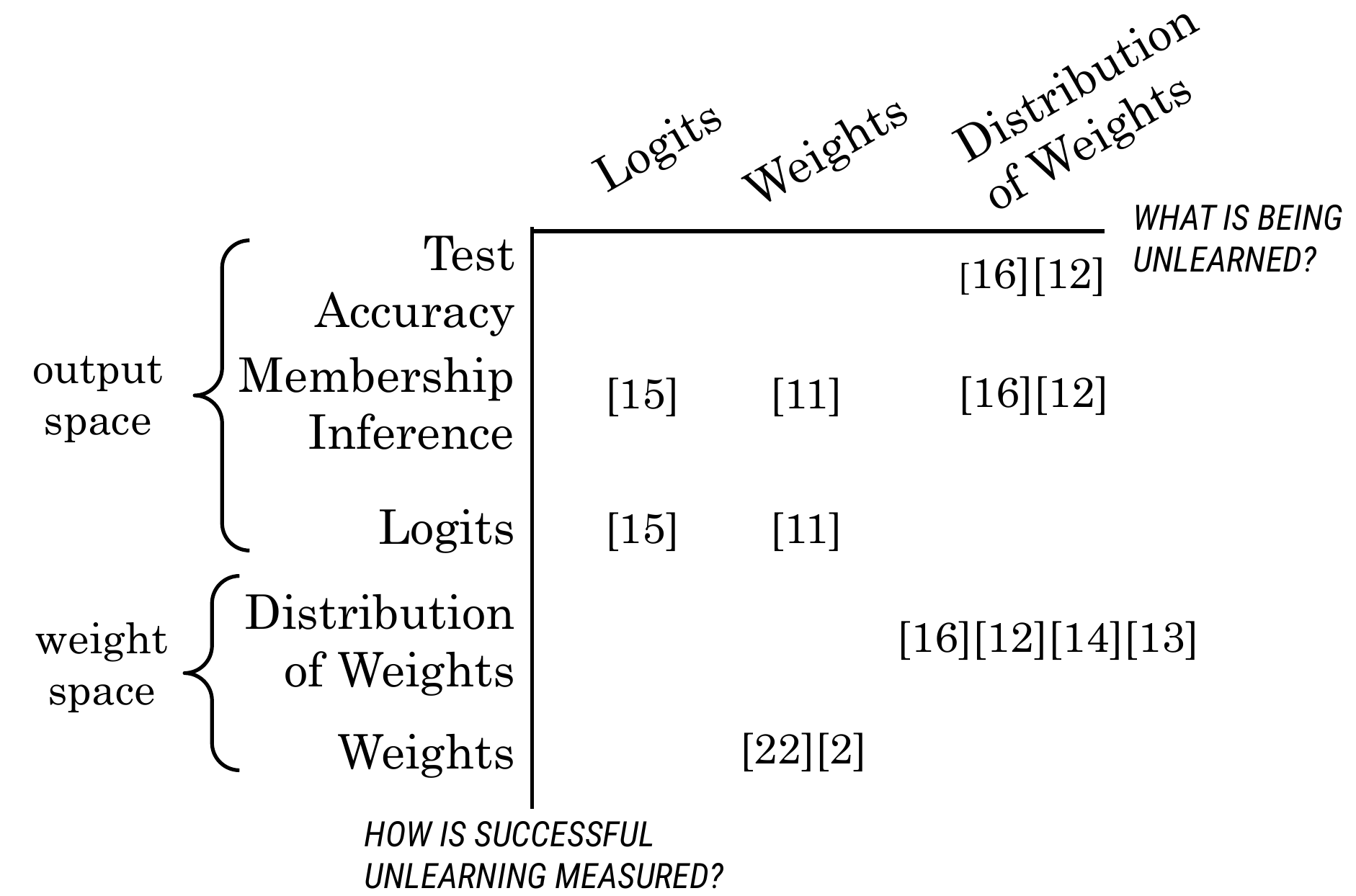}
    \caption{\small 
    Taxonomy of prior work on post-hoc (post-training) approximate (avoiding retraining) unlearning methods. Unlearning methods are categorized in two ways: (1) What is the definition of unlearning used to motivate the removal of information (horizontal axis)? (2) How is the success of the unlearning method measured (vertical axis)?}
    \label{fig:unlearning_table}
\end{figure}

We begin by discussing the semantics behind unlearning: the problem of {\em forgetting} a datapoint $\mathbf{x}^* \in D$ from a model $M$ which was  trained on it (\S~\ref{sec:define_unlearning}). We proceed to discuss metrics to measure the efficacy of unlearning (\S~\ref{sec:unlearning_metrics}), and methods to achieve them (\S~\ref{sec:unlearning_methods}). Broadly speaking we follow the taxonomy given in Figure \ref{fig:unlearning_table}, with emphasis on specific examples of the metrics (y-axis) and methods (x-axis).


\subsection{Defining Unlearning}
\label{sec:define_unlearning}

Let $\mathcal{H}_{D}$ define the distribution of all models a training rule could return when trained on a dataset $D$. $\mathcal{H}_{D}$ is a distribution and not a single point as SGD has inherent stochasticity and there is randomness with back-end floating point operations. Similarly let $\mathcal{H}_{D'}$ represent the distribution of all the models the same training rule returns when trained on dataset $D'= D \setminus \mathbf{x}^*$ where $\mathbf{x}^* \in D$ is the datapoint that is to be unlearned. Lastly, let $U(M,\mathbf{x}^*)$ be some process (randomized or deterministic) that takes a model $M \sim \mathcal{H}_{D}$ and a data point $\mathbf{x}^*$, and returns another model $M''$. Now if $\mathcal{S} = U(\mathcal{H}_D, \mathbf{x}^*)$ is the distribution of $\mathcal{H}_D$ after the transformation by $U$, we say the process $U$ is an {\em exact unlearning process} iff $\mathcal{S} = \mathcal{H}_{D'}$, \ie the distribution of output models from $U(\mathcal{H}_{D},\mathbf{x}^*)$ is the same distribution as $\mathcal{H}_{D'}$. In this case, $M'' \sim U(M,\mathbf{x}^*)$ is called an unlearned model. 

Though this definition looks precise, there is ambiguity in what metric space these models belong to (and consequently for the distributions). A model can be viewed either as just a mapping of inputs to outputs $(\mathbf{x},M(\mathbf{x}))$ in which case $\mathcal{H}_D$, $\mathcal{S}$, $\mathcal{H}_{D'}$ are distributions over a function space (\ie continuous function with the supremum metric), or as the specific parameters $\weights$ for an architecture, in which case $\mathcal{H}_D$, $\mathcal{S}$, $\mathcal{H}_{D'}$ are distributions over the weight space (\eg some finite dimensional real vector space with the euclidean norm). The ambiguity leads to two notions of \textit{exact unlearning}.
\begin{enumerate}
\itemsep0em
\item [{\bf Def 1}] $d_{\weights}(\mathcal{S},\mathcal{H}_{D'}) = 0$ where $d_{\weights}$ measures difference in the \textit{distribution of weights}
\item [{\bf Def 2}] $d_{out}(\mathcal{S},\mathcal{H}_{D'}) = 0$ where $d_{out}$ measures difference in the distribution of input-output maps, or simply difference in the \textit{distribution of outputs}
\end{enumerate}

Naively retraining without $\mathbf{x}^*$ as the unlearning process $U$ guarantees both of these definitions (as we exactly obtain $\mathcal{H}_{D'}$), and is the undisputed baseline for exact unlearning. However the issue with naive retraining is the large computational overhead associated with it (\ie cost to train a model), especially for large models~\cite{brown2020language}. 

\textit{Approximate unlearning} methods try to alleviate these cost-related concerns. Instead of retraining, these methods execute computationally less expensive operations on the final weights~\cite{guo2019certified,graves2020amnesiac,sekhari2021remember},  apply some architectural change~\cite{baumhauer2020machine}, or filter outputs~\cite{baumhauer2020machine}. Approximate unlearning also relaxes the definition of {\bf Def 1} or {\bf Def 2} to requiring the distance between $\mathcal{H}_{D'}$ and $\mathcal{S}$ to be small rather than being exactly zero.

Observe that there can be many different metrics $d$ over both of these spaces. Since comparing distributions is expensive (especially if it involves running the training algorithm multiple times to obtain multiple samples), approximate unlearning methods often move away from requiring a measure of the distance $d$ between distributions to instead a measure of an alternative quantity that captures the difference in the weight space or output space on a point basis. For example, a popular approach is to use membership inference~\cite{shokri2017membership} to gauge how close two models (before and after unlearning) are in the output space. The popular classes of metrics used are listed on the y-axis of our taxonomy in Figure~\ref{fig:unlearning_table}, and what aspect of the model the approximate unlearning methods change to unlearn a datapoint are listed on the x-axis.

\subsection{Unlearning Metrics} 
\label{sec:unlearning_metrics}
\vspace{1mm}

We now provide some examples of the various unlearning metrics used in past work. This follows the main categories found on the y-axis of Figure~\ref{fig:unlearning_table}. Note the first three are metrics over the weight space (\ie the metrics take as inputs weights or distributions of weights), and the fourth is in the output space (\ie metrics take as inputs the outputs of the model).

\vspace{1mm}
\noindent{{\bf 1. $\ell_2$ Distance}:} Here, we compute the weights of a naively retrained model $M'$ and compare them to the weights of the model $M''$ that we obtained after using an approximate unlearning process. If they are close, then this suggests that $M''$ is close to an exactly unlearned model, and can approximate the unlearned model. The standard approach to measure distance in the weight space is the $\ell_2$ distance; we term this {\em verification error} throughout the paper. Such an approach is used by Wu \etal~\cite{deltagrad}. 



Despite this metric being simple, there are certain drawbacks. First,  to compute the verification error, one first needs to compute $M'$ (through naive retraining), which is computationally expensive. If one could obtain $M'$, then they could avoid approximate unlearning altogether and use that as the unlearned model. Another issue is that one can retrain on the same sequence of data from the same initialization and obtain different terminal weights~\cite{jia2021proof} (which is caused by randomness and numerical instabilities in floating point operations).

\vspace{1mm}
\noindent{{\bf 2. KL Divergence}:} To bypass this issue of hardware randomness in training, Golatkar \etal~\cite{golatkar2020eternal} consider measuring the similarity of the weights of $M''$ with respect to a distribution of weights of $M'$. They achieve this by looking at {\em Kullback-Leiber (KL) divergence} of the two distributions (as they consider $M''$ as being the result of another stochastic process on the original model $M$). The drawback to this are that explicitly calculating the KL divergence requires knowing the final distributions of models trained on $D'$. This in turn involves sampling many final models trained on $D'$ in order to fit some distribution to the known distribution of $M'$. 


\vspace{1mm}
\noindent{{\bf 3. Privacy Leakage of Weight Distributions}:} Another common metric for measuring unlearning that also looks at the distribution of weights is the {\em privacy leakage from the distribution of weights of $M''$} (\eg~\cite{dwork2014algorithmic}): ideally, the distribution of weights of $M'$ leaks no information about $\mathbf{x}^*$ as it was not trained on it (though this is not necessarily the case~\cite{gupta2021adaptive}), and so the smaller the privacy leakage of $\mathbf{x}^*$ from the distribution of weights of $M''$, the closer $M''$ is to being unlearned. Sekhari \etal~\cite{sekhari2021remember} analogously present privacy leakage for unlearning in the framework of differential privacy (DP): they bound the leakage of information about a particular $\mathbf{x}^*$ (or more generally a set of datapoints to unlearn) with privacy parameters $\varepsilon,\delta$ and define this as $(\varepsilon,\delta)$-unlearning. Guo \etal~\cite{guo2019certified} work with a similar setup to Sekhari \etal, though not identical as they consider only the $\varepsilon$ bound part. In both cases, having (or decreasing) a bound on privacy leakage is presented as unlearning better.

\vspace{1mm}
\noindent{\bf 4. Membership Inference:} A privacy attack like membership inference (MI) only requires access to the model's predictions (outputs) to determine whether the unlearned model $M''$ was or was not trained on the point to be unlearned $\mathbf{x}^*$.
This approach is used by Graves \etal~\cite{graves2020amnesiac} and Baumhauer \etal~\cite{baumhauer2020machine}. The reasoning here is that as $M'$ was properly retrained without $\mathbf{x}^*$, a MI attack should return that it was not trained with $\mathbf{x}^*$; if a MI attack on $M''$ also consistently outputs that $M''$ was not trained with $\mathbf{x}^*$, then one could argue that their outputs are similar. However, such approaches are not \textit{precise}: it is possible a MI attack will give a false positive that $M'$ had indeed trained on $\mathbf{x}^*$, in which case simply lowering the MI likelihood of $\mathbf{x}^*$ for $M''$ does not necessarily imply it has unlearned. We revisit the limitations of MI in \S~\ref{ssec:MI_connection}.

\subsection{Unlearning Methods}
\label{sec:unlearning_methods}

We turn our attention to the following question: \textit{how exactly does one unlearn}? Current approaches are broadly categorized on the x-axis of Figure~\ref{fig:unlearning_table}.




\vspace{1mm}
\noindent{\bf Logits:} Baumhauer \etal~\cite{baumhauer2020machine} consider a specific type of model taking the form $M(\mathbf{x}) = \weights f(\mathbf{x})$, where $f$ is some black-box feature extractor, and $\weights$ is some parameterized matrix which is what is modified by training. They then unlearn a class from the linear layer {\em only}. This unlearning is done by defining a filtration matrix and appending it to the model which shifts the outputs to what one would get if one did not train with that class.

\vspace{1mm}
\noindent{\bf Weights:} Graves~\etal~\cite{graves2020amnesiac} proposed amnesiac machine learning, which is an unlearning protocol that logs all the training updates. Then, upon receiving an unlearning request for $\mathbf{x}^*$, one proceeds to add back all the training updates that involved $\mathbf{x}^*$ to the final weights to obtain an unlearnt model $M''$.

\vspace{1mm}
\noindent{\bf Distribution of Weights:} The scrubbing procedure introduced by Golatkar \etal~\cite{golatkar2020eternal,scrubbingv2} is derived by adding a weighted term to the loss (to measure the KL divergence of the distribution of final weights), and minimizing the added term. 
Alternatively, DP provides bounds on how indistinguishable a given model is to one not trained on any one of its training datapoints, in essence stating that a model has already unlearned. However Sekhari \etal~\cite{sekhari2021remember} show that differentially private learning only allows one to delete $O(\frac{n}{\sqrt{d}})$ data points while retaining meaningful bounds, and propose an unlearning algorithm for (strongly) convex loss functions that improves the number of datapoints one can delete by adding a hessian update and noise to the final weights.

\textcolor{black}{
\section{Verification Error \& Other Metrics}
\label{sec:verification}
\label{sec: One Metric to Rule Them All}
The taxonomy in \S~\ref{sec:overview_unlearning} captures the different unlearning metrics, which in turn represent the different aspects of a model that can change with the training data. Understanding the relationship between these metrics is paramount to concretely understanding the unlearning guarantees they provide. To this end, we focus on understanding what metrics verification error can and can not capture. 
}


\textcolor{black}{
\subsection{Verification Error implies $L^p$ Weight Metrics}
\label{sec: ver_err_implies_weights}
We will prove that verification error provides a bound on the supremum norm (or $L^{\infty}$ norm\footnote{See \cite{folland1999real} for more on $L^p$ spaces.}) of the difference of the distribution of weights obtained by (a) naively retraining, and by (b) using an approximate unlearning method\footnote{\textcolor{black}{In our final result we will require the unlearning method to only depend on the initial weights and batch ordering.}}. Note that a bound on the $L^{\infty}$ norm implies bounds on all $L^p$ norms for $p\geq 1$ (refer Propostion 6.10 in \cite{folland1999real} and by definition, distributions are in $L^1$). Informally, this allows us to motivate verification error as a metric that bounds a large class of weight space metrics.} 


Similarly, a bound on the supremum norm also implies guarantees similar to DP (in particular $\varepsilon=0$ and $\delta = b$ where $b$ is the bound on the supremum norm) by the reverse triangle inequality. The opposite, though, is not true when $\varepsilon \neq 0$ (as the bound now depends on the magnitude of the function and thus is not uniform). However, it should be noted that in our subsequent analysis, we make several assumptions about the stochasticity of training which does not allow us to derive strict DP-like guarantees. \textcolor{black}{In particular, we will assume that noise in the final weights for a fixed training sequence is a function of only the number of steps, and changes negligibly when removing a single step from each epoch.} We explain why we made these assumptions later in the section during our derivation. \textcolor{black}{Similarly our analysis will focus on unlearning with batch size 1 for simplicity.}

In particular, the focus here will be first comparing the probability distributions of final weights obtained by training $m$ epochs starting at a fixed initial weight $\weights_0$ with dataset $D$ and dataset $D'= D \setminus \mathbf{x}^*$ with SGD (and not mini-batch SGD, \ie considering mini-batch sizes of 1). We will represent the density functions respectively as $\mathbb{P}(\weights)$ and $\mathbb{P'}(\weights)$ where both are functions of the weights of the model $\weights$. Note that in what follows, $\mathbf{x}^*$ is fixed but can be any datapoint (\ie is a constant).

Let the size of $D$ be $|D|=n$, and let $I = \{\mathbf{x}_{i_1},\mathbf{x}_{i_2},\cdots,\mathbf{x}_{i_{mn}}\}$ denote a particular ordering of the datapoints when training on $D$ for $m$ epochs (where $i_j \in [n])$. Note that we have $(n!)^m$ such combinations; for $D'$, we have $(n-1)!^m$ such combinations. For each $I$, there exists an $I'$ (corresponding to $D' = D \setminus \mathbf{x}^*)$, which is formed by removing all instances of $\mathbf{x}^*$ (in-place) from $I$. Note that for a given $I'$, we have $n^m$ corresponding choices of $I$ (as for each epoch we have $n$ choices to place $\mathbf{x}^*$ per epoch, and $m$ epochs in total). Let $\weights_I$ represent the deterministic final weights resulting from training with the data ordering $I$, starting from $\weights_0$. \textcolor{black}{Let $\mathbf{g}$ denote the random variable which represents the noise on the final weights when training for $m \cdot n$ steps.} Thus, the random variable representing the final weights when training with $I$ is $\mathbf{z}_{I} = \weights_I + \mathbf{g}$. Similarly, we define $\mathbf{z}_{I'} = \weights_{I'} + \mathbf{g}'$ where $\mathbf{g}'$ represents the noise when training $(n-1) \cdot m$ steps. Note that, if training is deterministic, we have $\weights_{I'} = \weights_{I} - \mathbf{d}_I$ where $\mathbf{d}_I$ is some constant that only depends on $\weights_0$ and the ordering $I$ (as under deterministic operations $\weights_{I'}$ and $\weights_{I}$ are constants depending only on $\weights_0$ and the ordering $I$ and thus $\weights_{I} - \weights_{I'} = \mathbf{d}_I$ is a constant that only depends on them). Lastly, we make an assumption that the noise from training with $m \cdot n$ steps is equivalent to that training from $(n-1) \cdot m$ steps, \ie that $\mathbf{g}$ = $\mathbf{g}'$. This is based on the fact that this noise should only be a function of the number of training steps, and that noise per step is small (see \cite[\S~6]{jia2021proof} for a discussion on noise during training)\footnote{Thus, adding or removing $m$ steps (1 per epoch) when compared to the $m \cdot n$ total steps when training with ordering $I$ is insignificant}. Thus we have $\mathbf{z}_{I'} = \mathbf{z}_{I} - \mathbf{d}_{I}$.

Given that SGD samples individual datapoints uniformly, the probability of obtaining a given data ordering is constant. Further, if $\mathbb{P}_I(\weights)$ is the density function corresponding to $\mathbf{z}_{I}$ and $\mathbb{P'}_{I'}(\weights)$ is the density function corresponding to $\mathbf{z}_{I'}$ (which is equal to $\mathbb{P}_{I}(\mathbf{w}-\mathbf{d}_{I})$), then $\mathbb{P}(\weights) = \frac{1}{n!^m}\sum_{I}\mathbb{P}_{I}$ and $\mathbb{P'}(\weights) = \frac{1}{(n-1)!^m}\sum_{I'}\mathbb{P'}_{I'}(\weights)$.

\begin{lemma}
If every $\mathbb{P}_I$ is Lipschitz with Lipschitz constant $L$ (which is true if $\mathbf{g}$ is gaussian), and if we let $d = \frac{1}{n!^m}\sum_{I} ||\mathbf{d}_I||_{2}$, then:
\begin{equation}
\label{eq: l2_prob_difference_1}
||\mathbb{P}(\weights) - \mathbb{P'}(\weights)||_{2} \leq L \cdot d
\end{equation}
\label{lem:dif_train_retrain}
\end{lemma}
We refer the reader to Appendix~\ref{sec: Appendix A: Proofs} for the detailed derivation. This result shows that by accounting for all the individual distributions (for each ordering $I$), we were able to obtain a Lipschitz condition on the combined distribution (of final weights obtained by training $m$ epochs starting at a fixed intial weight $\weights_0$).  The Lipschitz condition is expressed with respect to the datasets including or excluding the point to be unlearned.

Now, we are also interested in the difference between the density function obtained after the application of an approximate unlearning method on $M$ (\ie $\mathbb{P''}(\weights)$), and $\mathbb{P'}(\weights)$. If the approximate unlearning method only considers $\weights_0$, $I$ to define an unlearning update $\mathbf{u}_I$, and obtains the approximately unlearned weights by adding $\mathbf{u}_I$ to $\mathbf{w}_I$, \ie $\weights_{I}'' = \weights_I + \mathbf{u}_I$, then:

\begin{corollary}
If every $\mathbb{P}_I$ is lipschitz with constant $L$ (true if $\mathbf{g}$ is gaussian), and if we let $v = \frac{1}{n!^m}\sum_{I} ||\mathbf{d}_I + \mathbf{u}_I||_{2}$, then:
\begin{equation}
\label{eq: l2_dif_bounded_ver_err}
||\mathbb{P''}(\weights) - \mathbb{P'}(\weights)||_{2} \leq L \cdot v
\end{equation}
\label{cor:ver_err_bound_probs}
\end{corollary}
The detailed proof is in Appendix~\ref{sec: Appendix A: Proofs}. This naturally follows from the same accounting procedure used for the previous lemma, but now $\mathbf{d}_I$ have been swapped for $\mathbf{d}_I + \mathbf{u}_I$. This gives us a bound on the supremum (or $L^{\infty}$) norm by the average verification error $v$ (times some constant) on the difference of the probability distributions after approximate unlearning $\mathbb{P''}(\weights)$ and ideal retraining $\mathbb{P'}(\weights)$. Note, however, that there is a necessary assumption on the form of this approximate unlearning method: to obtain Equation~\ref{eq: l2_dif_bounded_ver_err}, we require the unlearning update to only be dependant on the initial weight $\weights_0$ and $I$. We will introduce such an unlearning approach in \S~\ref{sec:A Closed Form Approximation of SGD}. Finally, an analogous bound  going in the opposite direction is provided in Appendix~\ref{sec: Appendix A: Proofs}.

\textcolor{black}{
\subsection{Convergence in the Outputs over Finite Sets}
}
It is intuitive that having similar weights would mean having similar outputs, as clearly if the weights of two models are identical then so are their outputs. We can formalize this by looking specifically at the verification error defined as $v = ||\weights'' - \weights'||_2$ where $\weights''$ are the weights of $M''$ (the unlearned model) and $\weights'$ are the weights of $M'$ (the model unlearned by retraining). With $v$ defined as above, we have for any $\mathbf{x} \in D$ that $\lim_{v \to 0} ||M'(\mathbf{x}) - M''(\mathbf{x})||_{2} = 0$ under continuity of the outputs of the models as functions of the weights. Here, by continuity, we mean that fixing the input $\mathbf{x}$ to a model results in the output $M(\mathbf{x})$ being a continuous function of the weights. Thus, $\lim_{v \to 0}$ entails point-wise convergence of the outputs of $M''$ to the outputs of $M'$. Furthermore, note that smooth functions are locally lipschitz. Because DNNs are smooth functions, this relation in the limit is thus equivalent to a linear convergence on an upper bound on the $\ell_2$ difference between the two models' outputs. \textcolor{black}{Now note, considering only a finite set of inputs $\{\mathbf{x}_1,\cdots,\mathbf{x}_n\}$ (\ie a dataset), we can take the maximum local lipschitz constant of all $M(\mathbf{x}_i)$ as a function of its weights for a fixed input $\mathbf{x}_i$, and hence have linear convergence on the outputs for all points in the set (not just pointwise).} This means changes to the differences in weights lead to proportional changes to an upper bound on the $\ell_2$ difference in the models' outputs \textcolor{black}{over a dataset}. 


\textcolor{black}{However the opposite does not hold. Consider the following counterexample: if one permutes the weights of a neural network, the model's outputs remain identical~\cite{permutation}. Yet, the verification error would not be $0$ after such a permutation, and so having all the outputs be the same does not entail having the same weights.}

\textcolor{black}{
\subsection{Connection to Membership Inference}
\label{ssec:MI_connection}
We now describe an apparent lack of a consistent relationship between verification error and MI-based metrics such as privacy risk score (or PRS)~\cite{song2021systematic} (which will be further supported by results in \S~\ref{ssec:effect_PRS}). Note that PRS was shown to accurately represent the confidence a datapoint was used during training \ie likelihood of MI. We implemented PRS by taking a shadow model trained only on half the training set, and allow the MI adversary to access half the test set. We then constructed the conditional probability distribution given in Equation 15 of~\cite{song2021systematic}, where we estimate the per label conditional probability by dividing the range of the entropy losses (Equation 8 in~\cite{song2021systematic}) for a given label into bins. Once the conditional distribution for training and test are constructed, we evaluate the PRS for a given point by computing Equation 13 of~\cite{song2021systematic}. We now proceed to} empirically show how a decrease in PRS {\em does \textcolor{black}{not}} result in a decrease in verification error. 

\begin{figure}
    \centering
    \includegraphics[width = 3.2in]{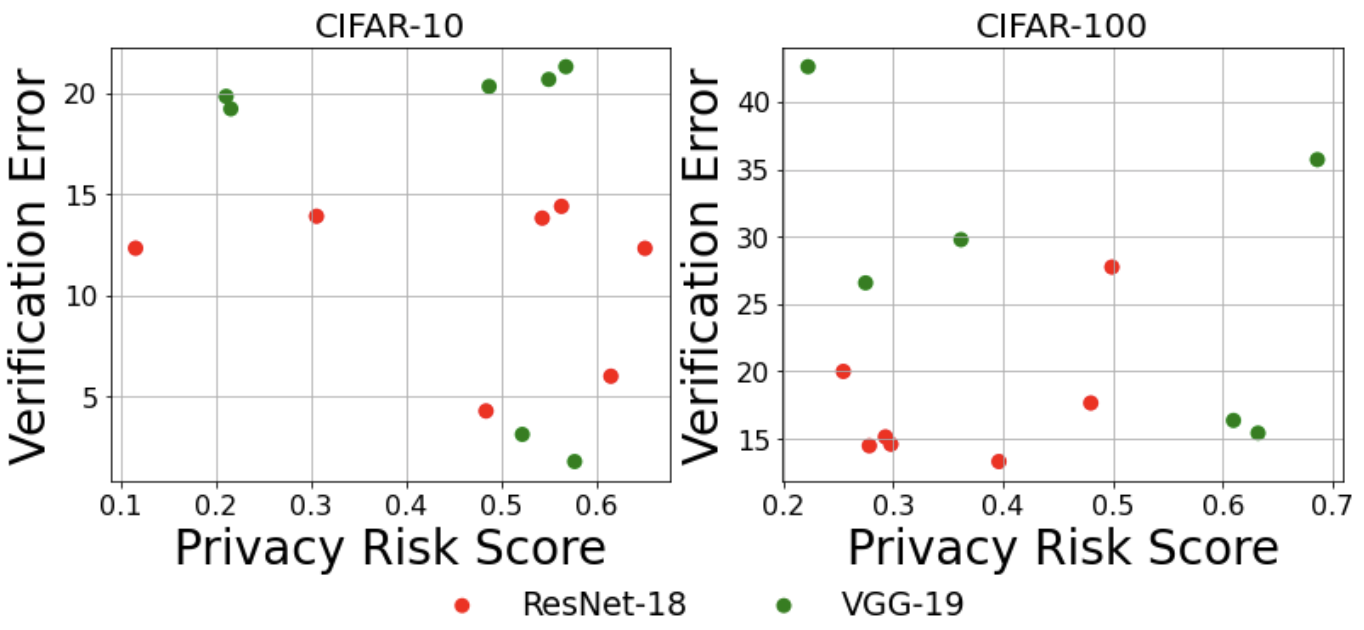}
 \caption{\small Correlation between privacy risk score and verification error in different training setups for CIFAR-10 and CIFAR-100. The correlations are -0.29 and -0.02, respectively.}
 \label{fig:amnesiac_prs_vs_ver}
\end{figure}

Specifically, we test the correlation between PRS and verification error after applying an approximate unlearning method to a datapoint $\mathbf{x}^*$. We choose amnesiac machine learning~\cite{graves2020amnesiac} because it was introduced with a MI-based criterion.\footnote{We reproduced the approach because it was originally evaluated on a MI attack by Yeom \etal~\cite{yeom2018privacy}.} We train the original model $M$, the naively retrained model $M'$, and  model $M''$ unlearned using amnesiac unlearning on the CIFAR-10~\cite{krizhevsky2009learning} and CIFAR-100 datasets~\cite{krizhevsky2009learning}, using the ResNet-18 and VGG-19 architectures~\cite{resnet, simonyan2014very}. We obtain 27 triplets of $M$,$M''$,$M'$ by changing the training settings (\ie different amounts of training, batch sizes, learning rates etc.). For each triplet, we measured the PRS of the unlearned point $\mathbf{x}^*$ on $M''$, and the verification error $||\weights'' - \weights'||_{2}$ (obtained from $M''$ and $M'$ respectively). The results are in Figure~\ref{fig:amnesiac_prs_vs_ver}. 

\textcolor{black}{The main takeaway is there is no monotonic relation between PRS and verification error. However it may be that under certain constraints (on the model, loss, training algorithm, etc.) there is a monotonic or even linear relationship between verification error and PRS, and future work may look into studying this. }


\section{Defining the Unlearning Error}
\label{sec:A Closed Form Approximation of SGD}

Recall that verification error's strongest limitation is the expense associated with calculating it (\ie the requirement for naive retraining). To circumvent this issue, we set out to approximate verification error. Our analysis seeks to understand the deterministic impact of a datapoint $\mathbf{x}^*$ on the final weights of a model when trained with SGD starting at initial weights $\mathbf{w}_0$. We expand the recursive updates performed by SGD to isolate terms that can be easily unlearned (because they do not depend on the order of datapoints) from terms that are difficult to unlearn. We then obtain and analyze a closed form approximation of these terms. This results in (a) a proxy metric (which we term {\em unlearning error}) that captures the effects of the verification error, and (b) an inexpensive approximate unlearning method that only depends on $\mathbf{w}_0$ and so allows us to use the bounds presented in \S~\ref{sec:verification} \footnote{\textcolor{black}{Do note future methods may expand our theory to consider methods not only dependent on $\mathbf{w}_0$.}}. 





\subsection{Expanding SGD} 

To understand the impact of a datapoint $\mathbf{x}^*$ on the final weights, we need to expand a sequence of SGD updates. We begin with the definition of a single SGD learning update:
\begin{equation*}
\label{eq1:}
\weights_{1} = \weights_0 - \eta \frac{\partial \mathcal{L}}{\partial \weights}|_{\weights_0,\hat{\mathbf{x}}_0}
\end{equation*}
where $\mathbf{w}_0$ denotes the weights at step 0 and $\hat{\mathbf{x}}_0$ denotes the data sampled at step 0. Note, we make no constraints on what $\mathbf{w}_0$ is (that is where training starts from), \ie could start from a pre-trained model). We obtain $\weights_{2}$, the weight obtained at step 2, as follows:
\begin{equation*}
\weights_{2} = \weights_0 - \eta \frac{\partial \mathcal{L}}{\partial \weights}|_{\weights_0,\hat{\mathbf{x}}_0} - \eta \frac{\partial \mathcal{L}}{\partial \weights}|_{\weights_1,\hat{\mathbf{x}}_1}
\end{equation*}
\begin{equation*}
\implies \weights_{2} = \weights_0 - \eta \frac{\partial \mathcal{L}}{\partial \weights}|_{\weights_0,\hat{\mathbf{x}}_0} - \eta \frac{\partial \mathcal{L}}{\partial \weights}|_{\weights_0-\eta \frac{\partial \mathcal{L}}{\partial \weights}|_{\weights_0,\hat{\mathbf{x}}_0},\hat{\mathbf{x}}_1}
\end{equation*}
This can further be expanded and approximated as follows:
\begin{equation*}
\weights_{2} \approx \weights_0 - \eta (\frac{\partial \mathcal{L}}{\partial \weights}|_{\weights_0,\hat{\mathbf{x}}_0} + \frac{\partial \mathcal{L}}{\partial \weights}|_{\weights_0,\hat{\mathbf{x}}_1} + \frac{\partial^2 \mathcal{L}}{\partial^2 \weights}|_{\weights_0,\hat{\mathbf{x}}_1} (- \eta \frac{\partial \mathcal{L}}{\partial \weights}|_{\weights_0,\hat{\mathbf{x}}_0}))
\end{equation*}
Our end goal is informed by our analysis in \S~\ref{sec: One Metric to Rule Them All}: we {\em must} focus on an analysis of verification error in SGD that is only a function of the initial weights $\mathbf{w}_0$ and the ordering of batches of data $I$. To scale the above for a sequence of $t$ updates, notice that the hessian terms from the expansion recursively depend on each other. Thus we have the final approximation:
\begin{equation}
\weights_{t} \approx \weights_0 - \eta \sum_{i=0}^{t-1} \frac{\partial \mathcal{L}}{\partial \weights}|_{\weights_0,\hat{\mathbf{x}}_i} + \sum_{i=1}^{t-1} f(i) \label{eq:approx}
\end{equation}
where $f(i)$ is defined recursively as:
\begin{equation}
\label{eq: recursivedef}
f(i) = -\eta \frac{\partial^2 \mathcal{L}}{\partial^2 \weights}|_{\weights_0,\hat{\mathbf{x}}_i} (- \eta \sum_{j=0}^{i-1} \frac{\partial \mathcal{L}}{\partial \weights}|_{\weights_0,\hat{\mathbf{x}}_j} + \sum_{j=0}^{i-1} f(j))
\end{equation}
with $f(0) = 0$.

\vspace{1mm}
\noindent{\bf Understanding the approximation.} From Equation~\ref{eq:approx}, observe that one can represent the weights obtained after training $t$ updates as the sum of two sums. We now wish to understand how $\mathbf{x}^*$ affects this expression (and thus the approximate outcome of training). The terms in the \textit{first sum} are simply gradients taken with respect to the initial weights $\weights_0$ and $\mathbf{\hat{\mathbf{x}}_i}$ following the order we give data to the model. Notice, however, that the exact order does not matter as we are simply adding them. In this case, it is clear that the effect of $\mathbf{x}^*$ (provided at any step in training) on this first sum is a gradient (or gradients) computed with respect to $\weights_0$ and $\mathbf{x}^*$. 

\begin{tcolorbox}
{\bf Single gradient unlearning:} To reverse this effect and unlearn, we simply have to \textit{add back these gradients} which amounts to adding $\frac{\eta m}{b} \frac{\partial \mathcal{L}}{\partial \weights}|_{\weights_0,\mathbf{x}^*}$ to the final weights, where $\eta$ is the learning rate, $b$ is the batch size, and $m$ is the number of epochs (which is the number of copies of this gradient that are present in the first sum).
\end{tcolorbox}





The \textit{second sum} is more complex as the expression is recursive, and so there is an inherent dependence on order. Understanding this sum will be our focus in \S~\ref{sec:approximation}. One of the main takeaways is that to compute all the terms that contain a particular $\mathbf{x}^*$, one would have to spend at least as much compute as training. For this reason, {\em the process to unlearn the first sum will be our unlearning method}, and our analysis of the second sum (which we do not unlearn) will result in an {\em inexpensive proxy metric} for the verification error which makes clear key variables one should focus on to decrease verification error. 


\subsection{Approximating the Second Sum Series}
\label{sec:approximation}

The global structure of the second sum follows $\eta^2 \mathbf{c}_{t} + \eta^3 \mathbf{c}_{t-1} + ... +\eta^{2+t-1} \mathbf{c}_{1}$, where $\mathbf{c}_{i}$ represents some vector. Thus, we can focus primarily on the $\eta^2$ dependent term as, in practice, we often observe $\eta$ being a magnitude or more less than $1$, and so $\eta^2$ is the dominant term. Thus, the second sum can be approximated as:
\begin{equation}
\label{eq:error}
\eta^2 \mathbf{c}_{t} = \sum_{i=1}^{t-1} \eta^{2} \frac{\partial^2 \mathcal{L}}{\partial^2 \weights}|_{\weights_0,\hat{\mathbf{x}}_i}  \sum_{j=0}^{i-1} \frac{\partial \mathcal{L}}{\partial \weights}|_{\weights_0,\hat{\mathbf{x}}_j} 
\end{equation}
We can further approximate $\sum_{j=0}^{i-1} \frac{\partial \mathcal{L}}{\partial \weights}|_{\weights_0,\hat{\mathbf{x}}_j}$ by its expectation $\frac{i(\weights_{t}-\weights_{0})}{t}$. We further normalize the vector, and thus have:
\begin{equation}
\label{eq: Detailed_Aprox}
\eta^2 \mathbf{c}_{t} \approx \eta^{2} \frac{||\weights_{t}-\weights_{0}||_{2}}{t} \sum_{i=1}^{t-1} \frac{\partial^2 \mathcal{L}}{\partial^2 \weights}|_{\weights_0,\hat{\mathbf{x}}_i} \frac{\weights_{t}-\weights_{0}}{||\weights_{t}-\weights_{0}||_{2}} i
\end{equation}
By focusing on $||\mathbf{c}_{t}||_{2}$, note that 
$$||\frac{\partial^2 \mathcal{L}}{\partial^2 \textbf{w}}|_{\textbf{w}_0,\hat{\mathbf{x}}_i} \frac{\textbf{w}_{t}-\textbf{w}_{0}}{||\textbf{w}_{t}-\textbf{w}_{0}||_{2}}||_{2} \leq ||\frac{\partial^2 \mathcal{L}}{\partial^2 \textbf{w}}|_{\textbf{w}_0,\hat{\mathbf{x}}_i}||_{2}$$ 
Now note the definition of the induced $\ell_{2}$ norm of the hessian, which is simply the square root of the largest eigenvalue of the hessian \ie $\sigma_{1,i}$ the first singular value of the hessian on $\hat{\mathbf{x}}_{i}$. For sake of simplicity we consider $\sigma = \max(\sigma_{1,i})$, and thus have the following inequality:


\begin{equation}
||\eta^{2} \mathbf{c}_{t}||_{2} \leq \eta^{2} \frac{||\weights_{t}-\weights_{0}||_{2}}{t} \cdot  \sigma \cdot \frac{t(t - 1)}{2} \label{eq:error bound}
\end{equation}
From this, we can observe that the second sum linearly depends on $||\weights_{t}-\weights_{0}||_{2}$, $\sigma$, and $t$. $\sigma$ captures the non-linearity of the loss landscape from the hessian, and so we see that the more linear the loss landscape, the tighter this bound on the approximation. We also see from this that in the scenario $t=1$ or $\sigma = 0$, this bound, and thus the term is 0, reinforcing the notion that if the loss landscape was just linear we could focus solely on the first sum of gradients.

\vspace{1mm}
\noindent{\bf Counting terms.} Now, with an understanding of what the dominant terms in the second sum of Equation~\ref{eq:approx} are, we take a step back and ask whether it is possible to unlearn $\mathbf{x}^*$ from these. What we will do is count all the terms involving $\mathbf{x}^*$; we then show that it is computationally expensive to compute those terms in order to remove their influence (and thus we can not reasonably improve our single-gradient unlearning method). Specifically, we focus on the expression in Equation~\ref{eq:error}. We let $i^*$ denote the first index where $\mathbf{x}^*$ appears, and we are interested in those terms that contain $\hat{\mathbf{x}}_{i^*}$ (and thus $\mathbf{x}^*$). These are the terms that we would need to remove in order to forget $\hat{\mathbf{x}}_{i^*}$ from the dominant terms in our second sum in Equation~\ref{eq:approx}. Note then that this will actually be an undercount, as $\mathbf{x}^*$ could appear in another batch; the final cost we state will be less than what we actually need to completely forget $\mathbf{x}^*$.


The first thing to note is that $\hat{\mathbf{x}}_i^*$ does not appear in any terms in the expression until $i= i^*$. For this index, we have $i^*$ terms dependent on $\hat{\mathbf{x}}_{i^*}$ as every term in the second sum in Equation~\ref{eq:error} is multiplied by $\frac{\partial^2 \mathcal{L}}{\partial^2 \weights}|_{\weights_0,\hat{\mathbf{x}}_{i^*}}$. Now for all the indices $i>i^*$ (\ie $t-1-i^*$ indices), we have exactly one term with $\hat{\mathbf{x}}_{i^*}$ which comes from a $\frac{\partial \mathcal{L}}{\partial \weights}|_{\weights_0,\hat{\mathbf{x}}_{i^*}}$ appearing in the second sum. In total, we have $(t-1-i^*)+i^* = t-1$ terms, and note that all these terms are hessian vector products (because all the terms in Equation~\ref{eq:error} are hessian vector products). In general, it is possible to implement hessian products in $O(n)$~\cite{pearlmutter1994fast}. 



Importantly, hessian vector products are at least as expensive as a gradient computation. When batch size $b=1$, this would be equivalent to the cost for training, and in general is $\frac{1}{b}\times$ the cost of training as we only need to compute $1$ of the $b$ gradients per batch, \ie just with respect to $\mathbf{x}^*$. However, this can still be a significant amount of computational expense in practice (as $t$ here is not the number of epochs but the number of individual training steps), and so from counting the terms with $\mathbf{x}^*$ we see that this second sum in Equation~\ref{eq:approx} (which we can approximate with Equation~\ref{eq:error}) is a bound to how well we can  reasonably forget the effect of $\mathbf{x}^*$.



\subsection{Unlearning Error}


Based on the above discussion for estimating the second sum, and how it bounds the efficacy of unlearning (without adding significant computational costs), we define Equation~\ref{eq:error bound} to be our \textit{unlearning error} ($e$). In practice, we make a slight modification as this bound can be loose; we take $\sigma$ to be the average of all the $\sigma_{1,i}$ rather than the $\max$, as to better approximate Equation~\ref{eq:error}. To be precise, setting $\sigma_{avg} = \frac{1}{t}\sum_{i=1}^{t} \sigma_{1,i}$, we define unlearning error as:
\begin{tcolorbox}
\begin{equation}
e = \eta^{2} \cdot  \frac{||\weights_{t}-\weights_{0}||_{2}}{t} \cdot \sigma_{avg}  \cdot \frac{t^{2} - t}{2} \label{eq:unl_err}
\end{equation} 
\end{tcolorbox}
Working with the average also allows one to only compute $\sigma_{1,i}$ every couple of steps, saving costs. Lastly, computing unlearning error $e$ does not require us to have applied our unlearning method first; we can pre-emptively know what it will be.

\section{Reducing the Unlearning Error}
\label{sec:Reducing the Unlearning Error}




In \S~\ref{sec:A Closed Form Approximation of SGD}, we identified terms of SGD which force (post-training) approximate unlearning approaches to incur verification error. Our unlearning error captures these effects through the number of steps $t$, the average singular value $\sigma_{avg}$, and the difference between final and initial weights $||\weights_t - \weights_0||_2$. Our closed form approximation of SGD from the previous section leads to an unlearning method (see gray box in \S~\ref{sec:A Closed Form Approximation of SGD}) that only depends on the initial weights $\weights_0$.  We now introduce modifications to the training procedure, which are orthogonal to the unlearning method itself but reduce the values of the different variables contributing to unlearning error. This therefore yields a training algorithm which makes it {\em easier to later unlearn} with our method at a minimal verification error. 



\subsection{Strawman Approaches}
\label{sec:Strawman Results}


We first consider two\footnote{We also tried several other ideas at reducing unlearning error which we do not describe in great detail here. One such idea was trying to reduce $\sigma_{avg}$ by adding a regularization term with the sum of the diagonal entries of the Hessian of CE loss with respect to the logits squared (which are of the form $\{p_1(1-p_1), \cdots, p_c(1-p_c)\}$ where $p_i$ is the softmax output of logit $M(\mathbf{x})_i$), but as will be theme with this section, we saw no benefit.} strawman approaches, each trying to directly minimize one of three factors identified in the unlearning error formulation: the number of steps $t$ and the difference between final and initial weights $||\weights_t - \weights_0||_2$. 





\vspace{1mm}
\noindent{\bf 1. Training for less.} From Equation~\ref{eq:unl_err}, recall that unlearning error $e$ depends on number of steps $t$. A simple method to reduce $e$ would be to train for fewer steps. We verify empirically if training for less steps $t$ reduces $e$.

Our evaluation is performed on two canonical architectures for computer vision, ResNet-18~\cite{resnet} and VGG-19~\cite{simonyan2014very}, on two vision datasets, CIFAR-10~\cite{krizhevsky2009learning} and CIFAR-100~\cite{krizhevsky2009learning}. The datasets consist of 60,000 $32\times32\times3$ images that belong to 10 and 100 classes respectively. The classes include animals and objects. This leads to four settings (one for each model and dataset combination). For a dataset $D$, we denote by $M_{t}$ the model trained on $D$ for $t$ training steps starting from $M_{0}$; in the experiments we compute the unlearning error from $M_{0}$ to $M_{t}$, and record the intermediate values of unlearning error $e$.

\begin{figure}[t]
        \centering
    \includegraphics[width=0.9\columnwidth]{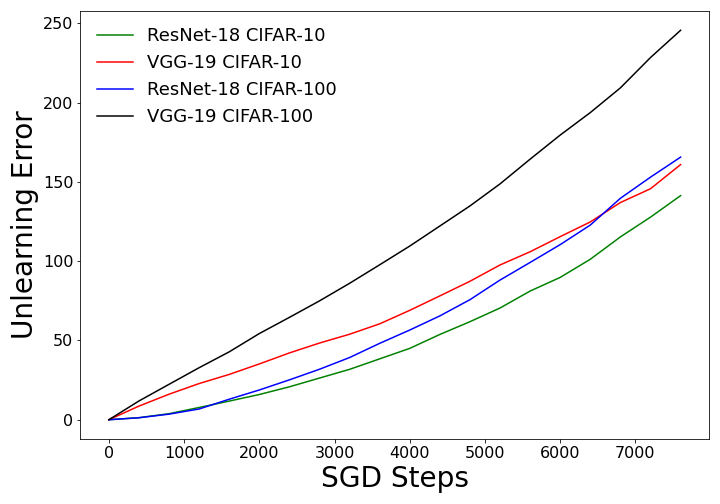}
    \caption{\small Unlearning error for 4 different settings as a function of the number of finetuning steps \textcolor{black}{with no pretraining}. Across all 4 settings, the unlearning error increases as a function of $t$. Each model was trained for $t = 7812$ steps (5 epochs)}
    \label{fig:unl_t}
\end{figure}

Figure~\ref{fig:unl_t} illustrates how $e$ increases  (practically linearly) with $t$. This implies that by training less, we decrease $e$ almost proportionally. This however also degrades the model's performance (\ie prediction accuracy) which creates an undesirable trade-off. One way to alleviate this limitation would be to pre-train the model on a public dataset so as to start training at a better initialization and compensate for the lower number of training steps performed on data that may be unlearned. However, it is not always possible to find a public dataset for which no unlearning requests will be issued. 




\vspace{1mm}
\noindent{\bf 2. $\ell_2$ regularization.} The next factor involved in the unlearning error $e$ is the weight difference $||\weights_t - \weights_0||_2$; decreasing this term (while not changing anything else) would also decrease unlearning error $e$. By the triangle inequality, we have that $||\weights_t - \weights_0||_2 \leq ||\weights_t||_2 + ||\weights_0||_2$ and so it would seem that by decreasing either $||\weights_t||_2$ or $||\weights_0||_2$, we might be able to obtain a tighter bound on $||\weights_t - \weights_0||_2$ and decrease it. This reasoning motivates using $\ell_2$ regularization which decreases the norm of the final weights $||\weights_t||_2$ by adding a $\lambda||w||_2$ regularization term (where $\lambda$ is a regularization constant) to the CE loss.

\begin{figure}[t]
  \centering
  \includegraphics[width=1.0\columnwidth]{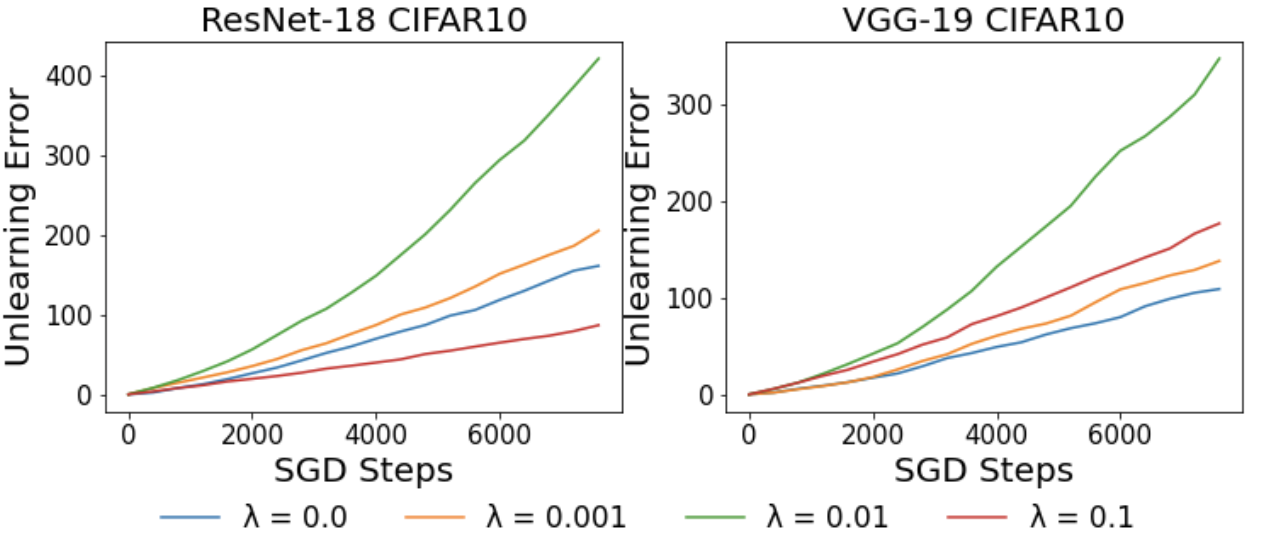}
 \caption{\small Unlearning error of 4 setups trained with increasing strength of  $\ell_2$ regularization. As shown, weight decay does not decrease the unlearning error consistently.}
\label{fig:l2_loss}
\end{figure}

To evaluate this regularizer, we use the same setup that was previously described. We repeat the experiments for different values of regularization strength: we pick $\lambda\in \{0.0, 0.001, 0.01, 0.1\}$. In Figure~\ref{fig:l2_loss}, we track $e$ for $t =7812$ training steps (5 epochs) for the two CIFAR-10 models (Table \ref{table: l2 reg} includes CIFAR-100 results) and the various strengths of $\ell_2$ regularization. We observe that there was no consistent benefit to employing the loss penalty, and in most cases it in fact increased $e$. The main issue is that the bound $||\weights_t - \weights_0||_2 \leq ||\weights_t||_2 + ||\weights_0||_2$ is very loose and we can decrease the right hand side without decreasing the left hand-side (the part which we are interested in) as seen in Figure~\ref{fig:delta_w_sigma_ell_2}. Thus, we next refine our analysis to obtain a tighter bound and a more effective regularization penalty. 




\subsection{Our Proposal: Standard Deviation Loss}
\label{ssec:std Loss}



In \S~\ref{sec:Strawman Results}, we considered two strawman approaches: decreasing training steps $t$ and $\ell_2$ regularization.
What we proceed to do in this section is one of the main contributions of this work: we introduce our own novel standard deviation regularizer. By way of studying a binary linear classifier, we show how it would (when training on an arbitrary, single data point) decrease the final change in weights by moving the minimum of the loss closer to the initial weights. Observe then that as long as this loss does not also increase the singular values (which we later empirically show in \S~\ref{sec:An Ablation Study}), it decreases the unlearning error $e$. Informally, the loss yields a model whose final weights required smaller gradient descent steps to get to from the initial weights. This explains why unlearning the contributions of a point to these steps using our single gradient approach leads to a smaller unlearning error when the model was trained with our loss.
This loss, which we will call {\em standard deviation (SD) loss}, is defined as follows: 

\begin{tcolorbox}
\small
\begin{multline}
\label{eq:SD_Loss}
\mathcal{L}_{SD}(M(\mathbf{x}),\mathbf{y}) \\ =
\mathcal{L}_{CE}(M(\mathbf{x}),\mathbf{y}) + 
\gamma~\sqrt{(\sum_{i=1}^{c}(M(\mathbf{x})_{i}-\mu)^{2})/c}
\end{multline}
\normalsize
\end{tcolorbox}
where $c$ is the output dimension of the model $M$, $\gamma$ is the regularization strength, and $\mu$ is the average value of the output logits $M(\mathbf{x})$ for a specific datapoint $\mathbf{x}$. 
Note that the SD loss is simple to integrate into a training framework as it just requires adding a regularization term to the loss function.

\subsubsection*{A Simple Binary classifier}
\label{sec:A Simple Binary Classifer}

Next, we illustrate our intuition behind the SD loss with  the case of a binary linear classifier (\ie a linear model with just two outputs) trained to minimize its loss on a single datapoint; we show how the SD loss reduces the minimum distance (in the weight space) to the minimum of the loss (which is where the final weights are found). This results in a lower weight difference  $||\mathbf{w}_t - \mathbf{w}_0||_{2}$, which in turn reduces unlearning error. This holds as long as the SD loss does not also simultaneously increase other variables involved in the definition of our unlearning error, \eg singular values. We confirm later in \S~\ref{sec:Evaluating SD loss} that this is the case (even for the neural networks we considered in our evaluation). 
\begin{figure*}[t!]
  \centering
 \subcaptionbox{$\gamma$ = 0.01 \label{fig21:a}}{\includegraphics[width=1.55in]{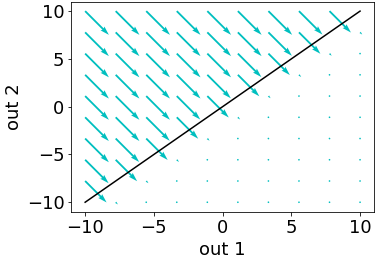}}\hspace{1em}%
   \subcaptionbox{$\gamma$ = 0.1 \label{fig21:b}}{\includegraphics[width=1.55in]{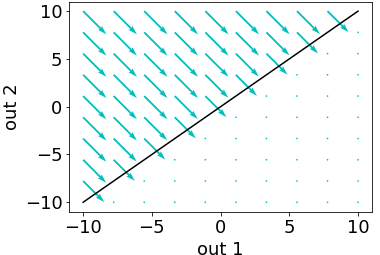}}\hspace{1em}%
   \subcaptionbox{$\gamma$ = 1 \label{fig21:c}}{\includegraphics[width=1.55in]{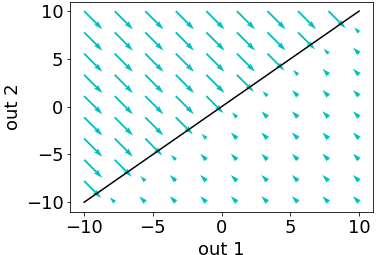}}\hspace{1em}%
   \subcaptionbox{$\gamma$ = 5 \label{fig21:d}}{\includegraphics[width=1.55in]{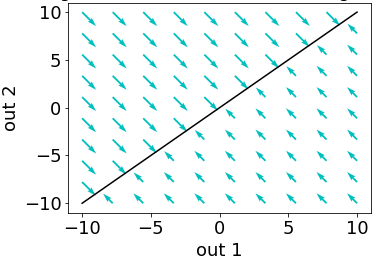}}
 \caption{\small Plots of the negative gradients of SD loss with respect to the two outputs ($a = out 1$, $b= out 2$). The black line represents $a=b$. Observe how the minimum of the loss landscape (where the arrows switch direction) approaches the black line.}
\label{fig:SD_loss_neg_grads}
\end{figure*}

Let us consider the specific setup where $\mathbf{x}$ is a vector of length $k$ representing a datapoint, $\weights = \begin{bmatrix} \weights_1 \\ \weights_2 \end{bmatrix}$ is a $2{\times}k$ weight matrix, and $M$ is a linear model with two outputs defined by $M(\mathbf{x}) = \weights\mathbf{x}$. For the sake of simplicity (and without loss of generality) let us assume the label of $\mathbf{x}$ is $0$. Then, the SD loss of our model, where $a = M(\mathbf{x})_1$ and $b = M(\mathbf{x})_2$ (\ie the first and second output), is:
\footnotesize
\begin{multline}
\label{eq:sd_loss_binary}
\mathcal{L}_{SD}(M(\mathbf{x}),0) = -\log(\frac{e^a}{e^a + e^b}) + \gamma~\sqrt{\frac{a^2+b^2-0.5(a+b)^2}{2}}
\end{multline}
\normalsize
where the second term is just a simplified version of the standard deviation.

\begin{figure}[t]
  \centering
 \subcaptionbox{$\gamma=0.01$ \label{fig22:a}}{\includegraphics[width=1.5in]{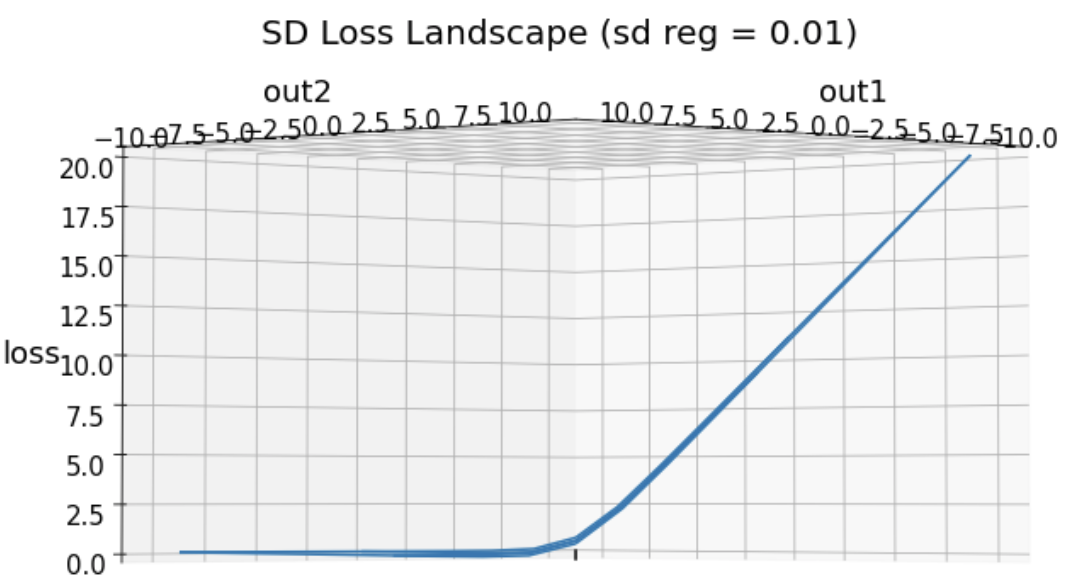}}\hspace{1em}%
   \subcaptionbox{$\gamma=0.1$ \label{fig22:b}}{\includegraphics[width=1.5in]{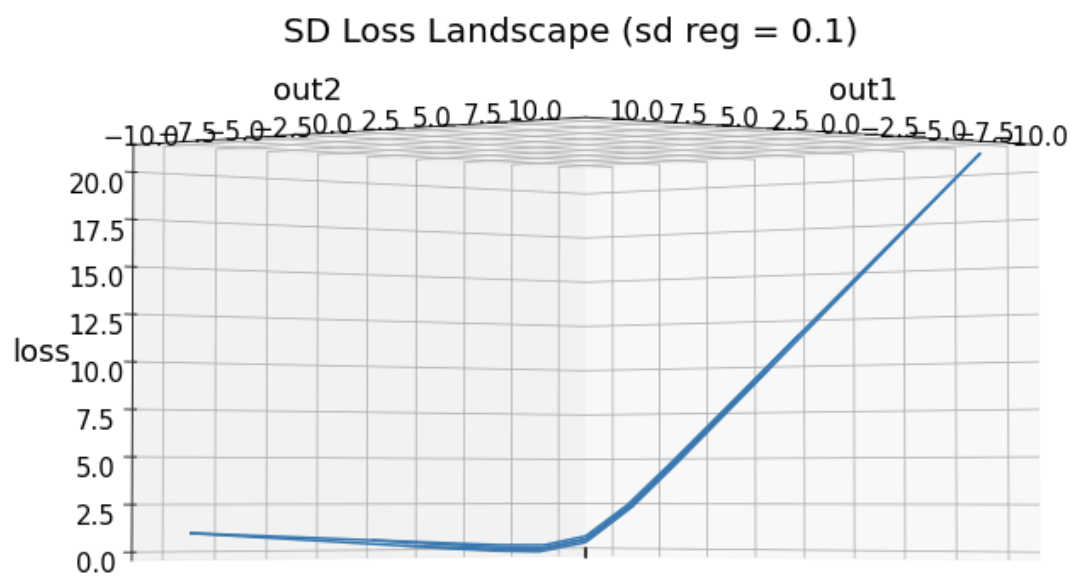}}\hspace{1em}%
   \subcaptionbox{$\gamma=1$ \label{fig22:c}}{\includegraphics[width=1.5in]{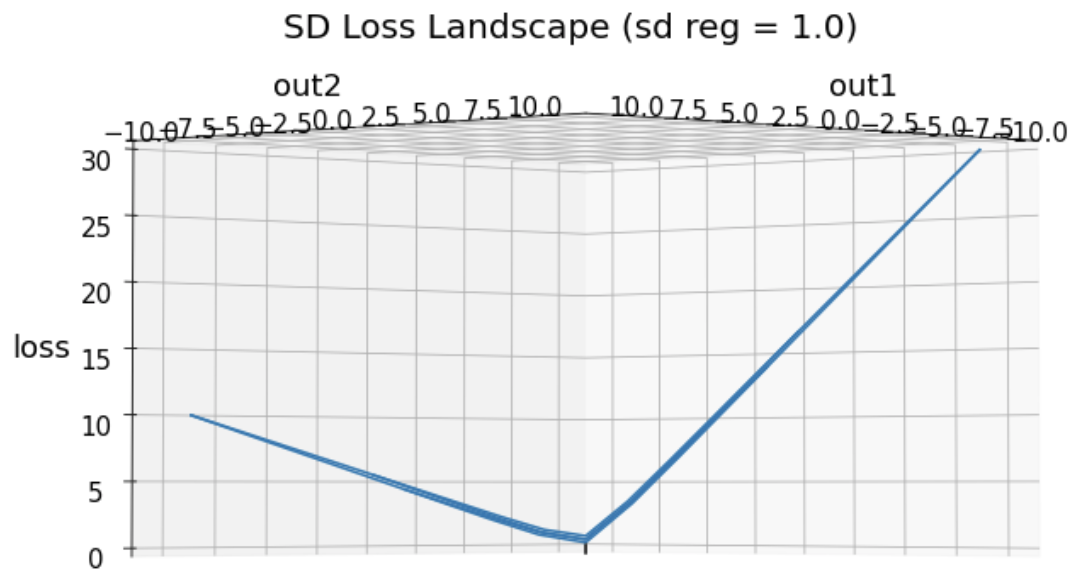}}\hspace{1em}%
   \subcaptionbox{$\gamma=5$ \label{fig22:d}}{\includegraphics[width=1.5in]{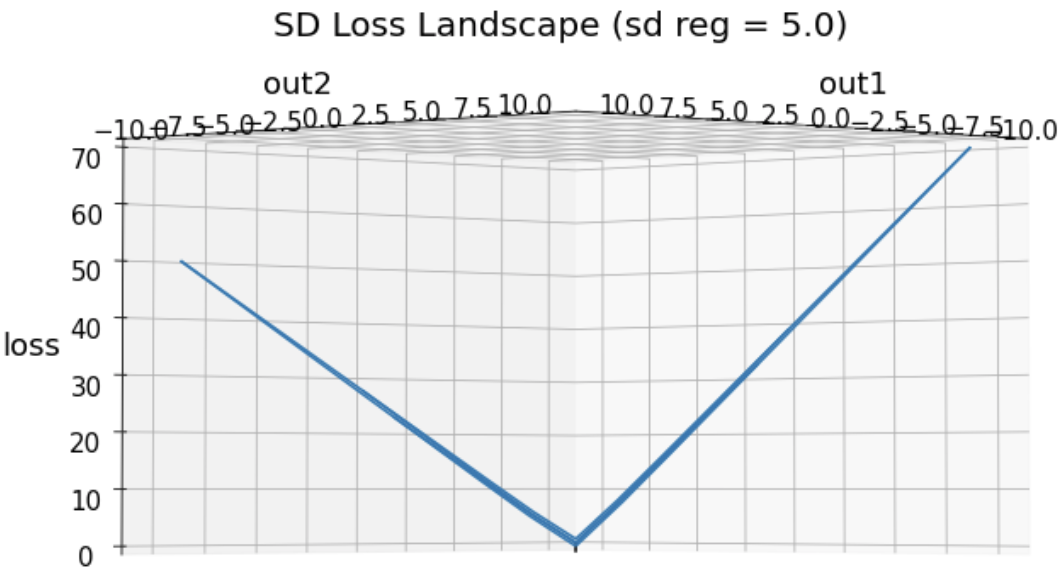}}\hspace{1em}%
 \caption{\small Plots of the SD loss landscape with respect to the two outputs ($a = out 1$, $b= out 2$). Note the black dot represents where $a=b$, and observe how the minimum of the loss landscape approaches the black dot}
\label{fig:SD_loss_landscape_1}
\end{figure}

The main empirical observation motivating the following analysis is the following: increasing the strength of $\gamma$ moves the minimum of the SD loss with respect to the outputs $a$ and $b$ (given by Equation~\ref{eq:sd_loss_binary}) closer to the line defined by $a=b$. This is illustrated in Figure \ref{fig:SD_loss_neg_grads} where we plot the gradients of our loss with respect to the outputs to observe where they are 0 (\ie when the direction flips). Similarly, Figure \ref{fig:SD_loss_landscape_1} illustrates how the minimum of the loss approaches the line $a=b$ (in black), and generally how the minimum is defined by a line $a=b + \epsilon$ where $\epsilon$ decreases with the strength of $\gamma$.

We now proceed to analytically show how SD loss reduces the minimum change in weights required to reach a minimum of the loss (for a binary linear classifier). We focus on the initialization of $\weights$ by a constant (that is all the entries are the same), and thus $a_0 = \weights_1 \cdot \mathbf{x} = \weights_2 \cdot \mathbf{x} = b_0$. To find the minimum distance in the weights space to a minimum of the loss, we need to solve the constrained optimization problem 


\begin{equation}\label{eq1}
  \begin{gathered}
    \min_{\mathbf{u}_1,\mathbf{u}_2} ||\mathbf{u}_1||_2^2 + ||\mathbf{u}_2||_2^2    \\
    \text{subject to} \\
    a_0 + \mathbf{u}_1 \cdot \mathbf{x} = b_0 + \mathbf{u}_2 \cdot \mathbf{x} + \epsilon \\
    \implies (\mathbf{u}_1 - \mathbf{u}_2) \cdot \mathbf{x} - \epsilon = 0
  \end{gathered}
\end{equation}

where $\mathbf{u}_1$ and $\mathbf{u}_2$ are the updates to $\weights_1$ and $\weights_2$ respectively to reach the minimum of the loss.

The solution, given by minimizing the lagrangian $\mathfrak{L} = \mathbf{u}_1 \cdot \mathbf{u}_1 + \mathbf{u}_2 \cdot \mathbf{u}_2 + \lambda((\mathbf{u}_1 - \mathbf{u}_2) \cdot \mathbf{x} - \epsilon)$, is $\mathbf{u}_1 = \frac{\epsilon}{2||\mathbf{x}||_2^2}\mathbf{x}$ and $\mathbf{u}_2 = \frac{-\epsilon}{2||\mathbf{x}||_2^2}\mathbf{x}$. Note that $||\mathbf{u}_1||_2^2 + ||\mathbf{u}_2||_2^2 = \frac{\epsilon^2}{2} ||\mathbf{x}||_2^2$, and by reducing $\epsilon$ we reduce the magnitude of the change of weights needed to reach a minimum of the loss. As increasing the strength of our SD loss regularization ($\gamma$) does just that, we see that increasing $\gamma$ means decreasing the minimum change in weights needed to reach a minimum (which is what the path following negative gradients approximates). Thus, if the SD loss does not (also) increase the singular values, it decreases the unlearning error. As a final remark, we showed that a state with $a = b$ is close to a minimum, but this implies a more general result of just being close to such a state where $a=b$ (\ie having low standard deviation in the outputs) means being close to the minimum.

\section{Implementation}
\label{sec: Implementation}

All of the experiments we describe next (and those performed in \S~\ref{sec:Strawman Results}) were conducted on T4 Nvidia GPUs with 16 GB of dedicated memory. We used Intel Xeon Silver 4110 CPUs with 8 cores each and 32GB of RAM. The underlying OS is Ubuntu 18.04.2 LTS 64 bit. We use \texttt{pytorch} v1.8.1 with CUDA 10.2 and \texttt{python 3.7}. We used the same datasets and models as described in \S~\ref{sec:Strawman Results}. We also evaluate our proposal on a different domain, {\em text}, using a pre-trained DistilBERT~\cite{sanh2019distilbert} fineunted on IMDb reviews~\cite{imdb} for sentiment analysis (a binary classification task); note IMDb has $25,000$ training and test datapoints consisting of sentences of varying lengths which we truncated to the first 512 tokens (the maximum DistilBert takes as input). \textcolor{black}{Note the size and variety of the datasets and models used in our evaluation is comparable or exceeds previous work in the topic of unlearning~\cite{deltagrad,golatkar2020eternal,guo2019certified}.}


\section{Evaluation}

Through our evaluation, we wish to answer the following questions. 
\begin{enumerate}
\itemsep0em
\item Does SD loss decrease unlearning error? 
\item How are the various components of the unlearning error ($t$, $\sigma_{avg}$, $||\weights_t - \mathbf{w}_0||_{2}$) affected by training with SD loss?
\item Does decreasing unlearning error (either using SD loss or changing $t$) result in a decrease in verification error? \textcolor{black}{Particularly, are they linearly related (strong Pearson correlation)?} 
\item \textcolor{black}{Can the effect of single gradient unlearning be measured by other metrics, such as PRS?}
\end{enumerate}


To answer the questions above, we utilize our single gradient unlearning approach to obtain the approximately unlearned model. Our salient results are: 

\begin{enumerate}
\itemsep0em
\item In \S~\ref{sec:Evaluating SD loss}, we conclude that the proposed SD loss is effective at decreasing the unlearning error. This is achieved at a moderate penalty to accuracy ($<3$ percentage points in the setups we tested).
\item In \S~\ref{sec:An Ablation Study}, we see that SD loss significantly decreases the final change in weights $||\weights_0 - \mathbf{w}_t||_{2}$, and that the impact of SD loss for a given regularization strength is concentrated at the beginning of training 
\item In \S~\ref{sec:Comparing Unlearning Error and Verification Error}, we see that decreasing unlearning error with any of the methods we have considered also decreases verification error (\ie they are strongly correlated) showing that unlearning error serves as a good proxy metric.
\item \textcolor{black}{In \S~\ref{ssec:effect_PRS}, we observe that single gradient unlearning decreases the PRS of the baseline (\ie $\gamma=0$). However, cases where $\gamma >0$ has no consistent effect.}
\end{enumerate}

\textcolor{black}{Note that while our theory is for batch size $b=1$, in particular Lemma \ref{lem:dif_train_retrain} and Corollary \ref{cor:ver_err_bound_probs}, our evaluation is for $b>1$ (where results are consistent across batch sizes). We believe future work may extend our theory for $b >1$.}




\subsection{SD Loss Decreases Unlearning Error}
\label{sec:Evaluating SD loss}
\begin{figure*}[t]
  \centering
  \includegraphics[width=7in]{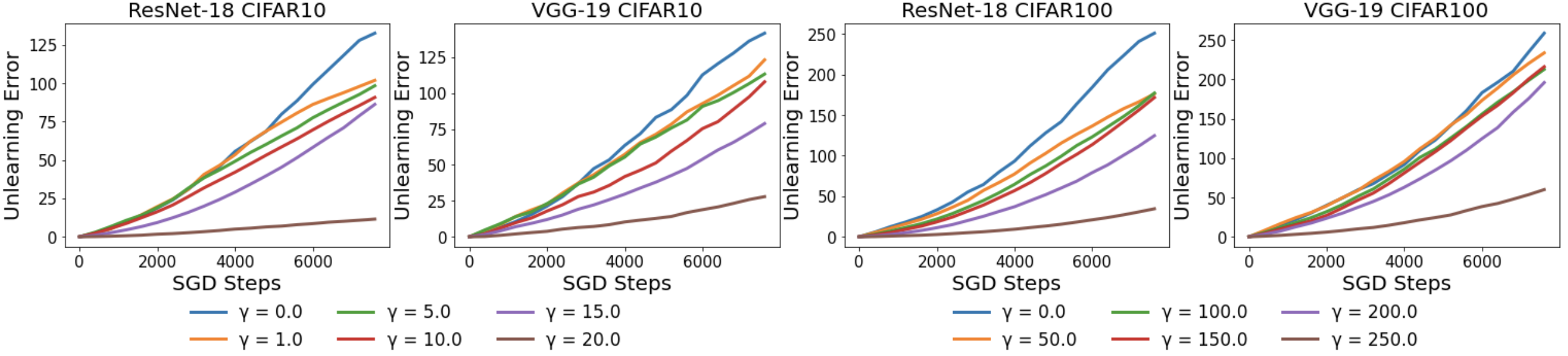}
 \caption{\small Unlearning error of 4 setups trained with the standard deviation loss with increasing levels of SD regularization. Across all 4 setups, a stronger regularization decreases the unlearning error at a (minimal) cost of the performance of the model.}
\label{fig:SD_loss}
\end{figure*}

Recall that we introduce the SD loss as a method to reduce the unlearning error as training commences. We wish to empirically validate this. We utilize the implementation setup details described in \S~\ref{sec: Implementation}, and the following setup:

\begin{enumerate}
\itemsep0em
\item Train $M_{0}$ on $D$ with SD loss for $N$ pre-training steps to obtain $M_{N}$.
\item Train $M_{N}$ for an additional $t$ steps on $\hat{\mathbf{x}}_1,\hat{\mathbf{x}}_2,\cdots,\hat{\mathbf{x}}_t$ (from $D$) to get $M_{N+t}$. 
\item Compute the unlearning error (Equation~\ref{eq:unl_err}) of $M_{N+t}$ starting from $M_N$, except here $\weights_t$ is the weights of $M_{N+t}$ and $\weights_{0}$ are the weights of $M_N$.
\end{enumerate}

In this setup, $N$ is a configurable parameter which allows us to understand how the amount of training affects the impact of SD loss on unlearning error (see \S~\ref{sec:An Ablation Study}). For the vision models, we utilize $N=\{0, 15625, 31250, 46875, 62500, 78125\}$ steps (\ie $0,10,20,30,40,50$ epochs, respectively). For the regularization strength, we use $\gamma = \{0, 1, 5, 10, 15, 20\}$ for CIFAR-10 and $\gamma = \{0, 50, 100, 150, 200, 250\}$  for CIFAR-100. In the case of DistilBERT fine-tuned on IMDB reviews, we observed we reached peak accuracy after just $N=4688$ steps (or $3$ epochs) and simply evaluated that for $\gamma = \{0.0, 0.01, 0.05, 0.1, 0.15, 1.0, 1.25, 1.5\}$ and discuss that independent of $t$.\footnote{$\gamma = 0$ means we just train with CE loss (which acts as our baseline)}



\noindent{\bf Results:} In Figure~\ref{fig:SD_loss}, we plot the unlearning error as a function of the number of training steps $t$. Observe that, as expected, the unlearning error grows as the training duration increases. However, increasing the regularization parameter $\gamma$ (from 0 \ie CE loss, to $>0$ values) reduces the rate or growth, and thus the final unlearning error. This is consistent with the final unlearning errors we observed for DistilBERT in Table~\ref{table:bert} where we see increasing regularization dropped final unlearning error by $40\times$. 

The influence of SD loss on the accuracy of the final (vision) models is presented in detail in Table~\ref{tb:SD_loss_unl_err} in Appendix~\ref{sec: Appendix C: Tables}; Table~\ref{tb:SD_loss_unl_err_10_cifar10} highlights results from our experiments with CIFAR-10 for $N=10$ epochs. Results for the DistilBERT experiments are in Table~\ref{table:bert}. For the vision models, note that as we increase the value of $\gamma$, the accuracy of the classifier learnt decreases. From Table~\ref{tb:SD_loss_unl_err_10_cifar10}, the decrease is within 5 percentage points for values of $\gamma \leq 5$. For values of $\gamma > 5$, the drop in accuracy is larger. This can be explained by having to fit to the dataset while also balancing the requirement of minimizing $||\mathbf{w}_t - \mathbf{w}_0||_2$. We will demonstrate this effect in more detail in \S~\ref{sec:An Ablation Study}. However, observe that the experiments with DistilBERT indicate a subtle increase in accuracy even after the regularization had cut the unlearning error by $2.5\times$, suggesting some domains/models can handle the regulariation better. More detailed analysis is required to explain this phenomenon. 
\begin{table}[ht!]
\resizebox{\linewidth}{!}{
\begin{tabular}{cSSSS}
\toprule
\textbf{Regularization}& 
\multicolumn{2}{c}{\textbf{ResNet-18}} & 
\multicolumn{2}{c}{\textbf{VGG-19}} \\ 
 \textbf{$(\gamma)$} & {$e$} & \textbf{Acc (\%) } & {$e$} & \textbf{Acc (\%) } \\ 
\toprule
 \multicolumn{1}{c}{0.0} & \multicolumn{1}{c}{132.49} &\multicolumn{1}{c}{77.11} & \multicolumn{1}{c}{141.73} &\multicolumn{1}{c}{77.69} \\ 
 \multicolumn{1}{c}{1.0} & \multicolumn{1}{c}{101.79} &\multicolumn{1}{c}{74.5} & \multicolumn{1}{c}{123.17} &\multicolumn{1}{c}{76.81}   \\ 
 \multicolumn{1}{c}{5.0} & \multicolumn{1}{c}{98.3} &\multicolumn{1}{c}{73.21} & \multicolumn{1}{c}{113.27} &\multicolumn{1}{c}{77.16} \\
 \multicolumn{1}{c}{10.0} & \multicolumn{1}{c}{90.71} &\multicolumn{1}{c}{69.13} & \multicolumn{1}{c}{107.9} &\multicolumn{1}{c}{74.07} \\
 \multicolumn{1}{c}{15.0} & \multicolumn{1}{c}{86.17} &\multicolumn{1}{c}{68.46} & \multicolumn{1}{c}{78.82} &\multicolumn{1}{c}{65.32} \\
 \multicolumn{1}{c}{20.0} & \multicolumn{1}{c}{11.5} &\multicolumn{1}{c}{34.74} & \multicolumn{1}{c}{27.94} &\multicolumn{1}{c}{31.18} \\
\bottomrule
\end{tabular}
}
\caption{\small Unlearning error ($e$) and testing accuracy on CIFAR-10 for a \textcolor{black}{pretraining amount of 10 epochs and $t$ equal to 1 epoch}.} 
\label{tb:SD_loss_unl_err_10_cifar10}
\end{table}

\begin{table}[ht]
\centering
\resizebox{\linewidth}{!}{
\begin{tabular}{cccc}
\toprule
{\bf Regularization ($\gamma$)} & {\bf $e$} & {$v$} & {\bf Acc (\%)} \\
\midrule
0 & 85.10 & 4.35 & 92.13\\
0.01 & 85.89 & 4.27 & 92.53\\
0.1 & 71.73 & 3.43 & 92.38\\
0.5 & 31.98 & 1.57 & 92.71\\
1.0 & 8.91 & 0.57 & 92.09\\
1.25 & 5.60 & 0.43 & 87.22\\
1.5 & 2.57 & 0.48 & 84.99\\
\bottomrule
\end{tabular}
}
\caption{\small Unlearning error $e$, verification error $v$ and testing accuracy for DistilBERT on the IMDB dataset for varying regularization amounts of the SD loss. \textcolor{black}{Note the pretraining amount is $3$ epochs and $t$ is 1 epoch.} }
\label{table:bert}
\end{table}




\subsection{An Ablation Study}
\label{sec:An Ablation Study}

\begin{figure}[t]
    \centering
    \includegraphics[width=1\columnwidth]{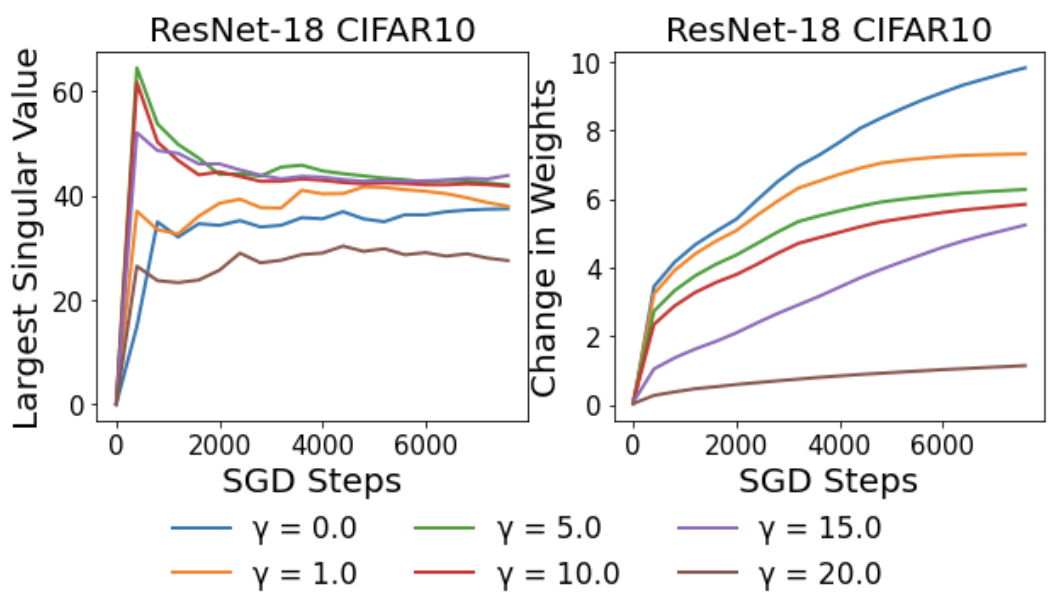}
    \caption{\small Largest singular value and change in weights over training for different regularization strengths for ResNet-18 on CIFAR-10. The change in weights decrease consistently and significantly as the regularization strength increases.}
    \label{fig:std_sigma}
\end{figure}

Recall that the unlearning error is dependent on $t$, $\sigma_{avg}$, $\mathbf{w}_t$, and also the value of $N$. As we can calculate the unlearning error with respect to different stages of training by varying the value of $N$, we can obtain a more nuanced view of how it evolves with respect to these factors. All results in this subsection are presented for the vision models only. The results are plotted for $N = 15,625$ steps (10 epochs), and $t = 7812$ steps (5 epochs) in Figure~\ref{fig:std_sigma}, but the trends and ensuing discussion hold for all other values of $N$. 

\noindent{\bf Singular values \& change in weights.} The impact of SD loss on $\sigma_{avg}$ and the final change in weights and largest singular values are presented in Figure~\ref{fig:std_sigma} (left). We see that the impact to the singular values is {\em minimal}. After some initial oscillation, SD loss leaves $\sigma_{avg}$ unchanged. Thus, the effect on reducing the unlearning error comes from elsewhere; we can see that SD loss {\em does decrease} the change in weights $||\mathbf{w}_t - \mathbf{w}_0||_{2}$ significantly with increasing $\gamma$\footnote{Here $w_t$ are the weights of the model $M_{N+t}$, and $w_0$ of the model $M_N$.}, as suggested by our analysis in \S~\ref{ssec:std Loss} (refer to Figure~\ref{fig:std_sigma} (right)). Recall that the change in weights influences the unlearning error, and smaller values of $||\mathbf{w}_t - \mathbf{w}_0||_2$ induces smaller values of the unleraning error.

\vspace{1mm}
\noindent{\bf Influence of $N$.} We wish to answer a more nuanced question: {\em how does the impact on SD loss vary as training progresses \ie as $N$ increases?} The results are presented in Table~\ref{tb:SD_loss_unl_err} in Appendix~\ref{sec: Appendix C: Tables}. We observe that as $N$ increases, the decrease in unlearning error caused by a given regularization diminishes (with the exception of ResNet-18 on CIFAR-10). Looking at VGG-19 on CIFAR-10 in Table~\ref{tb:SD_loss_unl_err}, we see w.r.t the {\em $N=10$ epochs of pre-training} row, $\gamma=5$ decreases the unlearning error by $20\%$ relative to the baseline ($\gamma=0$). With respect to the {\em $N=30$ epochs of pre-training} row, $\gamma=5$ results in a $4\%$ drop relative to the baseline ($\gamma=0$). This reduced effect is due to the model converging (or being close to converging) to its final weights: recall that in our setup, the weights obtained after $N$ steps of training are considered as $\mathbf{w}_0$; as $N$ increases, the value of $\mathbf{w}_t$ (which are the weights obtained after $N+t$ steps of training) are close to $\mathbf{w}_0$. 

\subsection{Unlearning Error \& Verification Error}
\label{sec:Comparing Unlearning Error and Verification Error}

Having established the relationship between unlearning error and SD loss, we wish to validate if unlearning error is a good proxy for verification error. To this end, we evaluate the relationship between unlearning error and verification error with a specific emphasis on training duration $t$, regularization strength $\gamma$, and a combination of the two datasets and model architectures. For this experiment, we utilize the setup (and notation) from \S~\ref{sec:Evaluating SD loss} (\ie both vision and text models) to compute unlearning error and the following setup to compute verification error:
\begin{enumerate}
\item Train $\model_{N}$ for $t-1$ steps on batches $\hat{\mathbf{x}}_2,\dots,\hat{\mathbf{x}}_t$. The resulting model is the naively retrained model $M'$ without batch $\hat{\mathbf{x}}_{1}$ (where note $\hat{\mathbf{x}}_{1}$ varies as we vary $N$) with weights $\weights'$.
\item Obtain the approximately unlearned model $M''$ (in particular $\weights''$) using our single gradient method as follows:
    $$ \weights'' = \weights_{N+t} + \frac{\eta}{b}\frac{\partial \mathcal{L}}{\partial \mathbf{w}}|_{\weights_{n+t},\hat{\mathbf{x}}_1}$$ where $\weights_{N+t}$ were the weights of $\model_{N+t}$.
\item Compute the verification error (\ie $||\weights' -\weights''||_{2}$).
\end{enumerate}

\begin{figure}[ht]
  \centering
 \includegraphics[width=0.99\columnwidth]{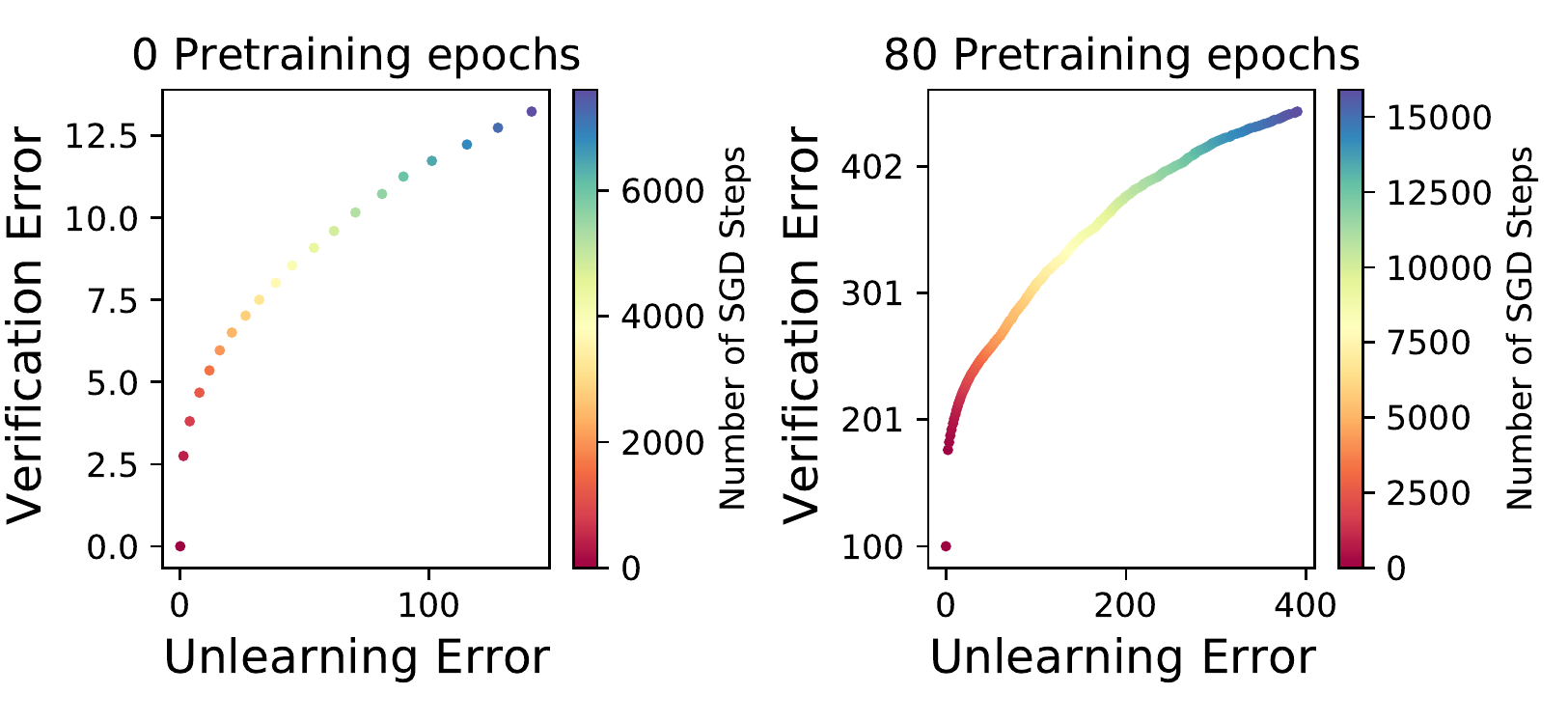}
 \caption{\small Unlearning and verification error as a function of the fine-tuning steps (indicated by the color gradient) for ResNet-18 on CIFAR-10 for $N=0$ steps, $ t =7,812$ steps (5 epochs) and $N= 250,000$ steps (80 epochs), $t = 15,625$ (5 epochs). Unlearning and verification error are measured every 400 and 100 SGD updates, respectively. The \textcolor{black}{Pearson} correlation between unlearning and verification error is {0.934} and {0.9668}, respectively. Note that \textcolor{black}{$t$ is the only variable giving a distribution.} Both unlearning error and verification error increase as a function of the number of fine-tuning steps.}
\label{fig:pretrain}
\end{figure}

\noindent{\bf Changing $t$:}\label{par:changin t} In Figure~\ref{fig:unl_t} in \S~\ref{sec:Strawman Results}, we saw that unlearning error increases with $t$. To this end, we utilize the ResNet-18 model trained using CIFAR-10. We wish to understand if verification error also increases with $t$. Figure~\ref{fig:pretrain} (left) shows the relationship between the verification error and the unlearning error as a function of the number of steps $t$ taken, for $t = 15,625$ (5 epochs) steps after $N=0$ steps. Observe that the two are strongly correlated (with a Pearson's coefficient of 0.934). When we increase the value of $N$ to $250,000$ steps (80 epochs) to measure how this effect is at a different stage of training (for the same value of $t$ as before), we observe the same effect: there is a strong correlation between verification and unlearning error (with a Pearson's coefficient of $0.9668$); refer Figure~\ref{fig:pretrain} (right). 



\begin{figure}[h!]
    \centering
    \includegraphics[width = 2.2in]{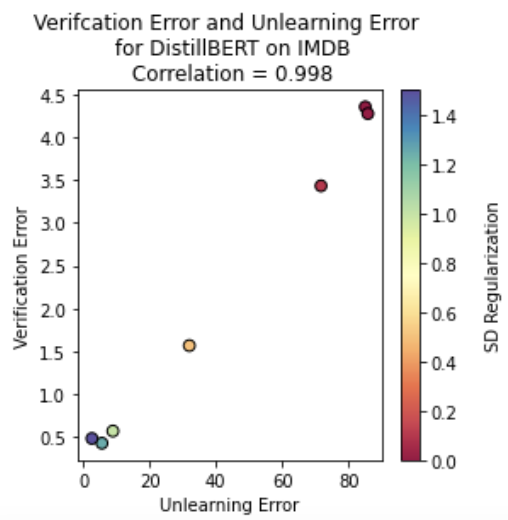}
    \caption{\small The \textcolor{black}{Pearson} correlation of verification error and unlearning error for DistilBERT on the IMDB dataset when varying the SD regularization strength.}
    \label{fig:correlation_DistilBERT}
\end{figure}

\begin{figure}[t]
    \centering
    \includegraphics[width=1.0\columnwidth]{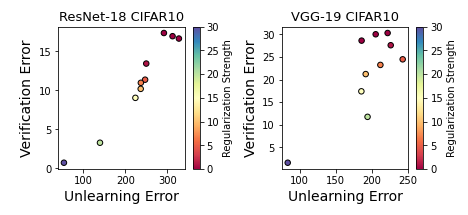}
 \caption{\small \textcolor{black}{Pearson} correlation between unlearning error and verification error for ResNet-18 and VGG-19 on CIFAR-10 \textcolor{black}{with regularization strength $\gamma$ being the only variable giving the distribution}. The correlations are both high at 0.96 and 0.81 for ResNet-18 and VGG-19, respectively. Notice that the stronger the regularization (more blue) the more the verification and unlearning error decrease in both settings.}
\label{fig:corr_std_loss}
\end{figure}

\vspace{1mm}
\noindent{\bf Changing $\gamma$:} Similarly, from Figure~\ref{fig:corr_std_loss}, we can observe that varying the strength of our regularization ($\gamma$) results in a strong \textcolor{black}{linear} correlation between unlearning error and the verification error. For ResNet-18 trained on CIFAR-10, the Pearson coefficient is 0.96, and the Pearson coefficient is 0.81 for VGG-19 trained on CIFAR-10. Furthermore for DistilBERT fine-tuned on IMDB reviews we observe a Pearson Coefficient of $0.998$ (see Figure~\ref{fig:correlation_DistilBERT}). Also notice that for large values of $\gamma$ (\ie the blue points near the origin), both verification and unlearning error are low. This further validates the efficacy of SD loss.




\begin{figure}[t]
    \centering
    \includegraphics[width=0.99\columnwidth]{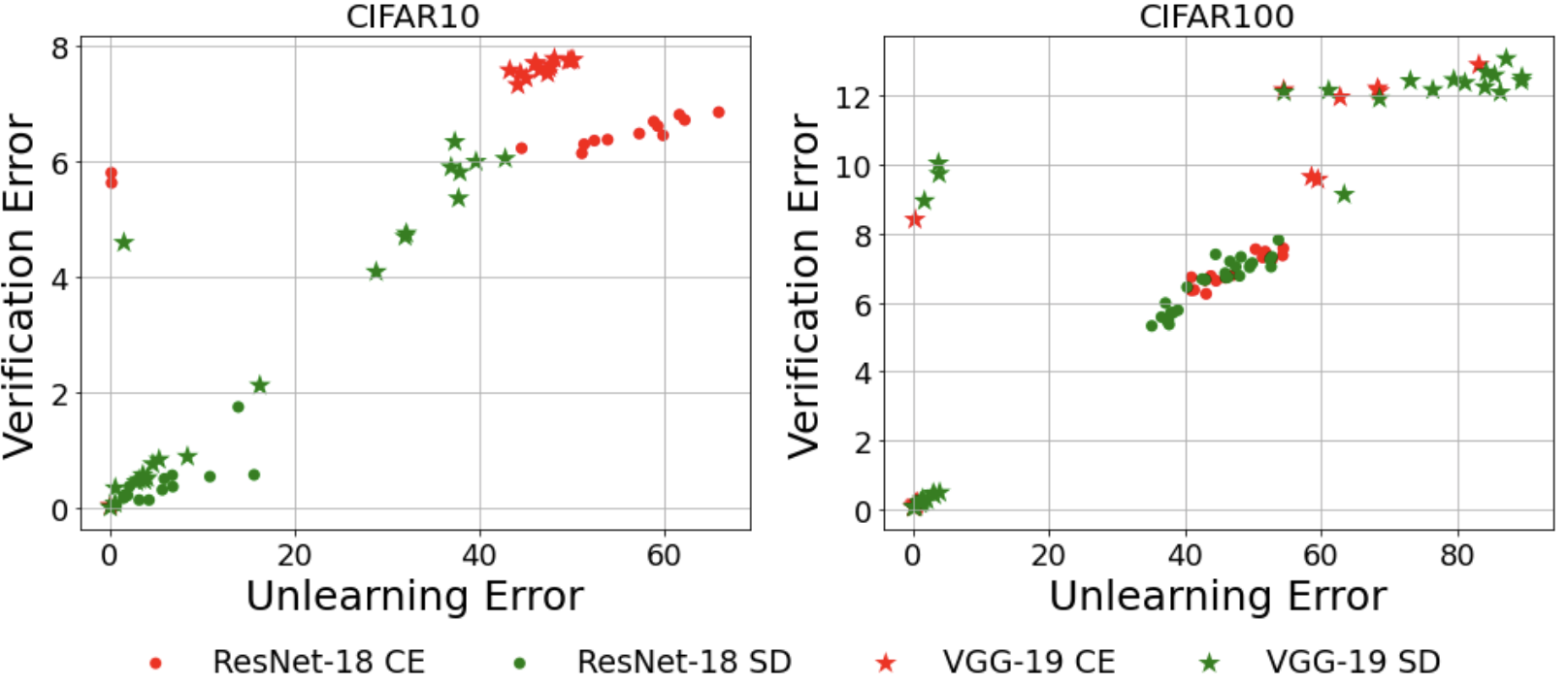}
 \caption{\small Unlearning and verification error for 160 settings across CIFAR-10 and CIFAR-100. The \textcolor{black}{Pearson} correlation between unlearning error and verification error for CIFAR-10 and CIFAR-100 are \textbf{0.9028} and  \textbf{0.8813}, respectively. Settings trained with SD loss (green) for both datasets have lower unlearning and verification error then those trained with regular CE (red).}
\label{fig:big_final_corr}
\end{figure}

\vspace{1mm}
\noindent{\bf Considering all models:} The goal now is to see if unlearning error can be used to compare which architectures and training setups are better at unlearning (\ie does unlearning error strongly correlate with verification error across training setups). This is seen in Figure~\ref{fig:big_final_corr} where we consider 160 different training setups for CIFAR-10 and CIFAR-100\footnote{We consider different amounts of pre-training $N = \{10,20, \dots, 80\}$ epochs, batch sizes $\{32,64,128\}$,  SD regularization strengths (values between the ranges shown in Figure~\ref{fig:SD_loss}), models (ResNet-18 and VGG-19) and batch size over which the hessian is calculated.}; we observe that generally when a model with a specific setup has lower unlearning error, it has lower verification error, and this relation is almost linear. 

This can be observed by looking at the strong positive \textcolor{black}{Pearson} correlation between unlearning error and the verification error across the different settings and architecture considered.

The fact that unlearning error is a good proxy for the verification error serves as evidence that the mathematical basis for unlearning error presented in \S~\ref{sec:approximation} holds for the different architectures, domains, and datasets we considered, despite the approximations we made to arrive at the definition of unlearning error. Another takeaway from this result, and the fact that the correlations stand across different architectures, is that unlearning error can be used to compare how well different models with potentially different architectures unlearn, leading to the possible study of what architectures unlearn better. We leave this aspect to future work.





\begin{figure*}[t]
  \centering
  \includegraphics[width=7in]{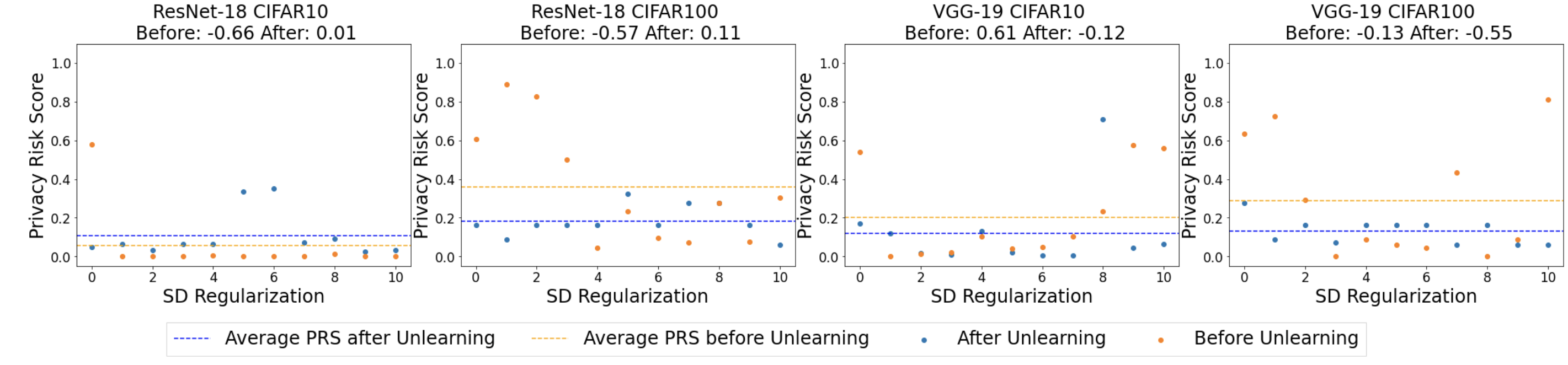}
 \caption{\small PRS of point to unlearn before and after unlearning as a function of the SD regularization trained with. Spearman correlation coefficients before and after unlearning are shown in the title of each of the plots. Note the inconsistencies in signs and magnitude of correlation before and after unlearning in all 4 settings.}
\label{fig:prs_after_before_unl}
\end{figure*}

\textcolor{black}{
\subsection{The Effect on PRS}
\label{ssec:effect_PRS}
We now study the effect of our single gradient unlearning method and training using SD loss on PRS, a metric which represents MI confidence. To do so, we calculate the PRS of the point to be unlearned before and after applying our unlearning method (when training was performed with varying strengths of SD regularization). The results are presented in Figure~\ref{fig:prs_after_before_unl}.
}

\textcolor{black}{
Observe that by applying single gradient unlearning on the baseline case (\ie $\gamma = 0$), we can consistently decrease the PRS by roughly $2\times$. On average, varying the regularizaion strength $\gamma$ decreases the PRS. However we find no consistent monotonic relation between SD regularization and PRS, as seen by the varying Spearman correlations (which switch sign and are often weak). Note we use Spearman here as we are asking if there is any monotonic relation, not necessarily linear. Regardless, the fact that SD regularization is not needed to decrease the PRS with our unlearning method, but greatly decreases verification error, highlights the dependence between seemingly independent methods.} 
\textcolor{black}{
\subsection{SISA Cost Comparison}
\label{ssec:SISA}
An interesting question is whether exact unlearning could ever be cheaper than approximate unlearning. As our method is a very cheap form of approximate unlearning, and SISA is the cheapest exact unlearning method, we compare how the costs of exactly unlearning with SISA~\cite{bourtoule2019machine} compares to our approximate unlearning costs. The cost of unlearning with SISA is at best $\frac{2S}{(R+1)} \times$ cost-to-retrain ($R$ is number-of-slices and $S$ is number-of-shards). Additionally, this requires $O(S.R)$ storage. Our cost is computing a single gradient (cost-to-retrain$/N$ where $N$ is number-of-steps) and requires no storage. When $\frac{2S}{(R+1)}<\frac{1}{N}$\, SISA could be faster, but then $R$ or $S$ is at least $\frac{\sqrt{N}}{2}-1$ (and typically $N \approx 100,000$). Thus $R$ and $S$ would be significantly higher than what was originally tested by Bourtoule \etal, and note that increasing $S$ was observed to decrease performance, but future work may investigate training with sufficiently large $S$ and $R$.
}

\section{Discussion}
\label{sec:Conclusion and Discussion}


We focus on discussing open questions raised by our and others' work.

\vspace{1mm}
\textcolor{black}{
\noindent{\bf When Is unlearning achieved:} The immediate question raised by our work is deciding when an entity has unlearnt. We see from the disparity between verification error and PRS that decreasing one does not necessarily mean decreasing the other. More over with verification error and unlearning error, we can reason about how much it has decreased from a baseline error (\ie the error with regularization $\gamma = 0$), but we do not have a scale of how much is enough. We believe answering when unlearning is achieved is an application oriented problem and depends, for instance, on what the user wants unlearning to achieve.
}

\vspace{1mm}
\noindent{\bf Experimental constraints:} We leverage the widely used implementation from \texttt{kaungliu} \footnote{\href{https://github.com/kuangliu/pytorch-cifar}{https://github.com/kuangliu/pytorch-cifar}} which achieves a $93\%$ test accuracy (on CIFAR-10) with ResNet18. We remove several enhancements, specifically:

\begin{enumerate}
    \item Learning rate ($\eta$) scheduler: our derivation assumes a constant LR.
    \item Data augmentation: introduces multiple copies of the unlearned data.
    \item Momentum/ADAM: as our approximation was focused on SGD.
\end{enumerate}

These restrictions are common when learning with differential privacy, and the drop to performance we experience on CIFAR-10 and CIFAR-100 is less significant. Nevertheless future work may be able to improve the performance by utilising other enhancements or dropping some of these restrictions. It is worth noting however that DistilBERT was still able to achieve high test accuracy. 


\vspace{1mm}

\begin{figure}[t]
    \centering
    \includegraphics[width = 0.99\columnwidth]{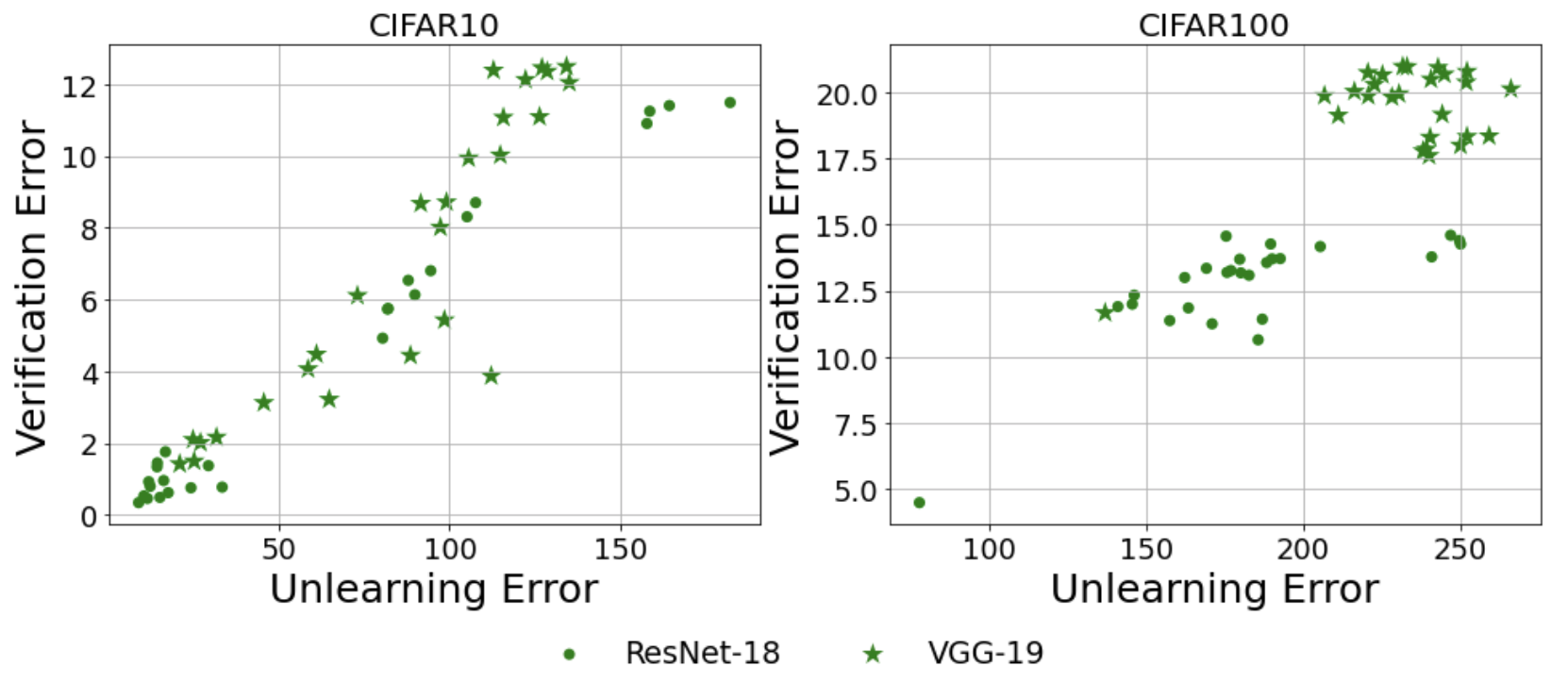}
 \caption{\small Unlearning and verification error for 104 settings across CIFAR-10 and CIFAR-100 using \textbf{amnesiac unlearning}. Correlations between unlearning error and verification error for CIFAR-10 and CIFAR-100 are resp. \textbf{0.91} and  \textbf{0.81}.}
\label{fig:big_final_corr_amnesiac}
\end{figure}

\begin{figure}[t]
    \centering
    \includegraphics[width = \columnwidth]{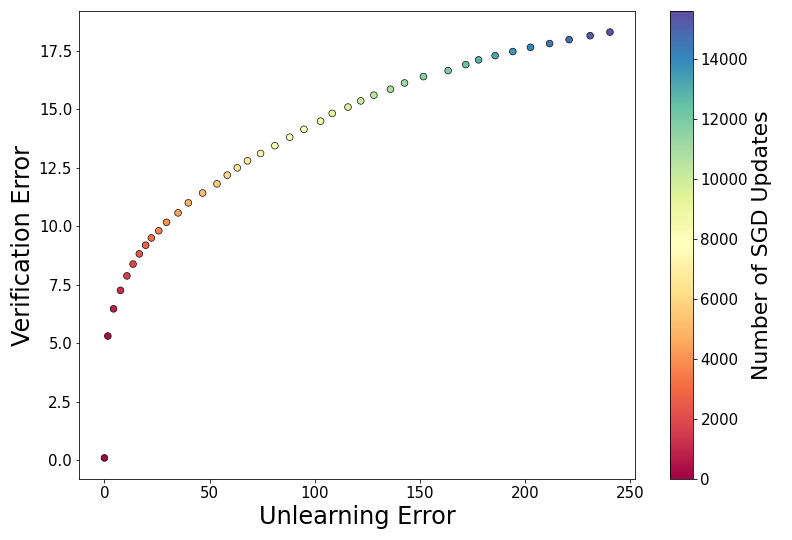}
    \caption{\small Unlearning error and verification error as a function of the finetuning steps (indicated by the color gradient) for ResNet-18 on CIFAR-10 using \textbf{amnesiac unlearning}}
    \label{fig:amnesiac_unl_vs_ver}
\end{figure}

\vspace{1mm}
\noindent{\bf Unlearning error and verification error for other unlearning methods:} An interesting question is whether our unlearning error serves as a good proxy for verification error even when employing unlearning methods besides our own. We focused on amnesiac machine learning \cite{graves2020amnesiac}, and computed the analogous result of Figure~\ref{fig:big_final_corr} in Figure~\ref{fig:big_final_corr_amnesiac} where we once again vary different batch sizes, training amounts, regularization strength(s), etc. except now  $M''$ is obtained via amnesiac machine learning. As was the case for our approach, we see very strong correlations between unlearning error and verification error of $0.91$ on CIFAR-10 and $0.81$ on CIFAR-100. \textcolor{black}{We similarly computed the analogous result of Figure~\ref{fig:pretrain} in Figure~\ref{fig:amnesiac_unl_vs_ver} and got comparable results to using our unlearning method.} This begs the question of whether unlearning error can be used to assess the verification error (by proxy) of  other approximate  methods for unlearning in SGD. 


\vspace{1mm}
\noindent{\bf Expanding analysis of verification error:} However, it should be noted that though amnesiac machine learning retains a strong correlation, it does not solely depend on $\weights_0$ or $I$. This is because it computes gradients with respect to intermediate weights during training; training from the same $\weights_0$ with the same batch ordering $I$, we get significantly different weights during the end of training. Thus, it violates the assumptions made by our analysis as it introduces significant noise we did not account for in our proofs. Thus the bounds presented in \S~\ref{sec: One Metric to Rule Them All} do not apply (\ie we are not sure if lowering verification error with these methods lowers the difference in probability distributions). However this does not mean similar bounds do not exist, and we leave it for future work to expand these results to other classes of unlearning methods and possibly expanding our existing assumptions. 

\vspace{1mm}
\noindent{\bf Architectures and unlearning error:} Recall from \S~\ref{sec:Evaluating SD loss}, that the results of the ResNet architecture on CIFAR-10 were atypical. The SD loss on ResNet-18 continued to significantly reduce unlearning error into the later stages of training and in general dropped unlearning error with a smaller cost to performance compared to VGG-19; we also see from Table \ref{tb:SD_loss_unl_err} that though SD loss effect diminishes for ResNet-18 on CIFAR-100, it is significantly less so than VGG-19. Similarly, DistilBERT in fact was able to drop unlearning error by a magnitude while seeing less significant drops to accuracy. This begs the question, are some models or domains just better for unlearning? We believe an interesting direction for future work is to pinpoint what architectural or domain features make unlearning easier or harder.

 
\vspace{1mm}
\noindent{\bf Regularizers and unlearning error:} Similarly future work might look into better regularizers, potentially ones focusing on reducing singular values, and improving the performance of our SD loss by a regularization scheduler that accounts for the degrading impact we noted in \S~\ref{sec:An Ablation Study}.

\section{Conclusion}


In this paper we first discussed past work on approximate unlearning, noting how there was a great breadth to what metric one can use to define unlearning. Following this we showed how verification error captures a large class of unlearning metrics (under some assumptions), motivating using it to define an approximate unlearning method. However, as verification error cannot be optimized directly when devising an unlearning method (as it requires a perfectly unlearned model for computation), we decomposed SGD with a Taylor series to first propose our single-gradient unlearning method, and furthermore propose our \textit{unlearning error} which we showed effectively proxies verification error when using our unlearning method. To then improve the effectiveness of our unlearning approach, we looked at the variables our proxy metric depends on and proposed our SD loss which we showed can effectively decrease the unlearning error (and thus the verification error) associated with our unlearning method. We expect that future work will try and improve the loss we used in our work, or see how it may help with different machine learning related tasks, and extend the bounds we presented in our work to other contexts and see if certain assumptions can be dropped.

\section*{Acknowledgements}

We would like to acknowledge our sponsors, who support our research with financial and in-kind contributions: CIFAR through the Canada CIFAR AI Chair program, DARPA through the GARD program, Intel, Meta, NFRF through an Exploration grant, and NSERC through the Discovery Grant and COHESA Strategic Alliance. Resources used in preparing this research were provided, in part, by the Province of Ontario, the Government of Canada through CIFAR, and companies sponsoring the Vector Institute. We would like to thank members of the CleverHans Lab for their feedback.

\section*{Availability}
Relevant code for the above experiments can be found at \url{https://github.com/cleverhans-lab/unrolling-sgd}.

\newpage
\bibliography{references}

\appendix

\subsection{Proofs}
\label{sec: Appendix A: Proofs}


\vspace{1mm}
\noindent{\bf Lemma 1}

We have:
\begin{multline}
||\mathbb{P}(\weights) - \mathbb{P'}(\weights)||_{2} = \\ 
||\mathbb{P}(\weights) - \mathbb{P'}(\weights) +  \frac{(n-1)!^m}{n!^m}(n^m\mathbb{P'}(\weights) - n^m\mathbb{P'}(\weights))||_{2} \\
\leq ||\frac{1}{n!^m}\sum_{I} \mathbb{P}_I(\weights) - \mathbb{P'}_{I'}(\weights)||_{2} + |(\frac{(n-1)!^mn^m}{n!^m}-1)|\mathbb{P'}(\weights) \\
\leq \frac{1}{n!^m}\sum_{I} ||\mathbb{P}_I(\weights) - \mathbb{P}_{I}(\mathbf{w}-\mathbf{d}_{I})||_{2}     
\end{multline}

Where the step from the first line to the second line follows from noting that by having $n^m$ copies of $(n-1)!^m\mathbb{P'}(\weights) = \sum_{I'}\mathbb{P'}_{I'}(\weights)$ we can associate to each $\mathbb{P}_I(\weights)$ it's corresponding $\mathbb{P'}_{I'}(\weights)$, which comes from the previous remark on counting of $I$ and $I'$. If we assume every $\mathbb{P}_I$ is lipschitz (which is true for Gaussian noise), then $||\mathbb{P}_I(\weights) - \mathbb{P}_I(\mathbf{w}-\mathbf{d}_{I})||_{2} \leq L_I||\mathbf{d}_I||_{2}$; note by definition all $\mathbb{P}_I$ are just the same distributions but shifted (as we have the same noise $\mathbf{g}$ for each plus some varying constant $\weights_I$), thus all the $L_I$ are equal, i.e., $L_I = L$ for every $I$ (as translating a function doesn't change its lipschitz constant). Lastly if we let $d$ be the average of all the $\mathbf{d}_I$, that is $d = \frac{1}{n!^m}\sum_{I} ||\mathbf{d}_I||_{2}$, we have

\begin{equation}
||\mathbb{P}(\weights) - \mathbb{P'}(\weights)||_{2} \leq Ld
\end{equation}


\vspace{1mm}
\noindent{\bf Corollary 1}

To understand how this relates to verification error and approximate unlearning, let us consider an approximate unlearning method which only looks at $\weights_0$ and $I$ and to $\weights_I$ obtains the approximately unlearned weights $\weights_{I}'' = \weights_I + u_I$ where $u_I$ is some unlearning update. Note then we can analogously define a $\mathbf{z}_{I}''= \weights_{I} + \mathbf{u}_I + \mathbf{g}$ as before with corresponding density function $\mathbb{P}''_{I}(w)$. Defining $\mathbf{v}_I = \mathbf{u}_I + \mathbf{d}_I$ we get $\mathbf{z}_{I'} = \mathbf{z}_{I}'' - \mathbf{v}_I$ where $\mathbf{v}_I$ represents the analytic (ignoring noise) verification error for the given ordering $I$ (i.e., the difference between the weights obtained from the approximate unlearning $\weights_{I}''$ and the retrained weights $\weights_{I'}$). At this point it is clear all the previous steps follow and defining $v$ as the average of all the $||\mathbf{v}_I||_2$ we get

\begin{equation}
||\mathbb{P''}(\weights) - \mathbb{P'}(\weights)||_{2} \leq Lv
\end{equation}

where now $\mathbb{P''}(\weights) = \frac{1}{n!^m}\sum_{I}\mathbb{P}''_{I}$ represents the probability density function of the weights after applying the approximate unlearning method on $M$ to obtain $M''$.
\vspace{1mm}
\noindent{\bf Reverse Direction of Corollary 1}
We now proceed to get a bound on the average verification error $v$ based on a uniform bound between the density function after the approximate unlearning and the density function for ideal retraining.

Note, assuming the noise has $0$ mean, $v$ is simply the difference in expectation of $\mathbb{P''}(\weights)$ and $\mathbb{P'}(\weights)$, and let us for the time being assume that the union of the support of $\mathbb{P''}(\weights)$ and $\mathbb{P'}(\weights)$ is bounded, i.e., if $\mathbf{W}$ is the union of the supports, then $\int_{\mathbf{W}} ||\weights||_{2} d\weights = a < \infty$. This is reasonable as we would expect the noise from training to be bounded; in fact in general we would only need the integral outside a bounded domain to be finite, but the reasoning is analogous to what follows and would simply add an extra constant term. Now if say $||\mathbb{P''}(\weights) - \mathbb{P'}(\weights)||_{2} < b$ for some scalar $b$ for all $w$ (i.e., $b$ is a uniform bound), we have:
\begin{multline}
\label{eq:exp_ver_err}
v = ||\mathbb{E}(\mathbb{P''}(\weights)) - \mathbb{E}(\mathbb{P'}(\weights))||_{2} \\
= ||\int_{\mathbf{W}} \mathbb{P''}(\weights) \weights d\weights - \int_{\mathbf{W}} \mathbb{P'}(\weights) \weights d\weights||_{2} \\
\leq \int_{\mathbf{W}} ||\mathbb{P''}(\weights) - \mathbb{P'}(\weights)||_{2} \cdot ||\weights||_{2} d\weights \\
\leq ba
\end{multline}
and so in this sense we see reducing the max value of $||\mathbb{P}(\weights) - \mathbb{P'}(\weights)||_{2}$ gives a smaller bound on the expectation of verification error $v$.



\subsection{Additional Figures}
\begin{figure}[H]
    \centering
    \includegraphics[width = 3.2in]{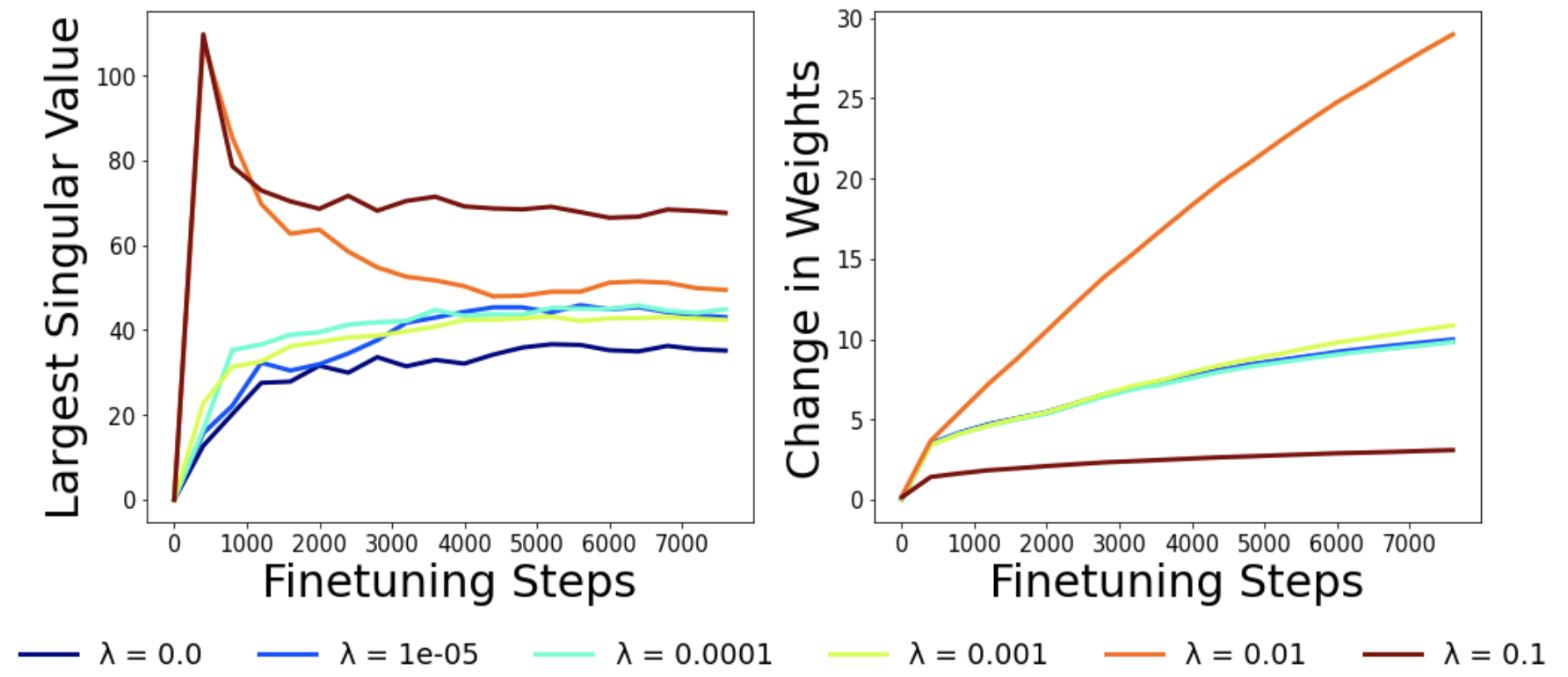}
 \caption{Change in weights and singular values during training for different strengths of $\ell_2$ regularization for ResNet-18 trained on CIFAR-10}
\label{fig:delta_w_sigma_ell_2}
\end{figure}

In Figure~\ref{fig:delta_w_sigma_ell_2}, the change in weights and singular values for varying strengths of the $\ell_2$ regularization is shown. 
Figure~\ref{fig:amnesiac_unl_vs_ver} and~\ref{fig:big_final_corr_amnesiac} are identical setups to Figure~\ref{fig:pretrain} and~\ref{fig:big_final_corr} but using amnesiac unlearning to produce the model $M^{''}$.


\subsection{Additional Tables}
\label{sec: Appendix C: Tables}
In Table~\ref{table: l2 reg}, the unlearning error and test accuracy for the 4 setups as a function of the $\ell_2$ regularization strength and pretraining amount (in epochs) are shown. Table~\ref{tb:SD_loss_unl_err} shows a similiar experiment but with the proposed SD regularization.

\begin{table*}[t]
\large
\resizebox{\linewidth}{!}{
\begin{tabular}{ccSSSScSSSS}
\toprule
\multirow{2}{*}{\textbf{Pretrain Epochs}} & 
\multirow{1}{*}{\textbf{Regularization}} & 
\multicolumn{2}{c}{\textbf{ResNet-18 CIFAR-10}} & 
\multicolumn{2}{c}{\textbf{VGG-19 CIFAR-10}} & 
\multirow{1}{*}{\textbf{Regularization}} & 
\multicolumn{2}{c}{\textbf{ResNet-18 CIFAR-100}} & 
\multicolumn{2}{c}{\textbf{VGG-19 CIFAR-100}} \\ 
 &\textbf{CIFAR-10} & \textbf{Unlearning Error} & \textbf{Accuracy (\%) } & \textbf{Unlearning Error} & \textbf{Accuracy(\%)} &\textbf{CIFAR-100} & \textbf{Unlearning Error} & \textbf{Accuracy(\%)} & \textbf{Unlearning Error} & \textbf{Accuracy(\%)} \\ 
\midrule 
\hfil\multirow{6}{*}{0}\hfill & \multicolumn{1}{c}{0.0} & \multicolumn{1}{c}{148.31} &\multicolumn{1}{c}{10.84} & \multicolumn{1}{c}{200.66} &\multicolumn{1}{c}{10.2} & \multicolumn{1}{c}{0.0} & \multicolumn{1}{c}{154.51} &\multicolumn{1}{c}{0.95} & \multicolumn{1}{c}{359.05} &\multicolumn{1}{c}{1.18}  \\ 
& \multicolumn{1}{c}{1e-05} & \multicolumn{1}{c}{145.44} &\multicolumn{1}{c}{10.68} & \multicolumn{1}{c}{174.23} &\multicolumn{1}{c}{10.87} & \multicolumn{1}{c}{1e-05} & \multicolumn{1}{c}{171.82} &\multicolumn{1}{c}{0.94} & \multicolumn{1}{c}{256.09} &\multicolumn{1}{c}{1.12}  \\ 
& \multicolumn{1}{c}{0.0001} & \multicolumn{1}{c}{164.48} &\multicolumn{1}{c}{8.78} & \multicolumn{1}{c}{165.01} &\multicolumn{1}{c}{10.53} & \multicolumn{1}{c}{0.0001} & \multicolumn{1}{c}{159.26} &\multicolumn{1}{c}{1.22} & \multicolumn{1}{c}{501.29} & \multicolumn{1}{c}{0.92}  \\ 
& \multicolumn{1}{c}{0.001} & \multicolumn{1}{c}{613.22} & \multicolumn{1}{c}{9.65} & \multicolumn{1}{c}{199.99} &\multicolumn{1}{c}{9.85} & \multicolumn{1}{c}{0.001} & \multicolumn{1}{c}{161.26} &\multicolumn{1}{c}{0.85} & \multicolumn{1}{c}{547.47} &\multicolumn{1}{c}{0.98}  \\ 
& \multicolumn{1}{c}{0.01} & \multicolumn{1}{c}{702.94} &\multicolumn{1}{c}{9.54} & \multicolumn{1}{c}{772.91} &\multicolumn{1}{c}{10.01} & \multicolumn{1}{c}{0.01} & \multicolumn{1}{c}{651.45} &\multicolumn{1}{c}{1.18} & \multicolumn{1}{c}{1162.51} &\multicolumn{1}{c}{1.0}  \\ 
& \multicolumn{1}{c}{0.1} & \multicolumn{1}{c}{2092.77} &\multicolumn{1}{c}{10.43} & \multicolumn{1}{c}{3980.42} &\multicolumn{1}{c}{10.52} & \multicolumn{1}{c}{0.1} & \multicolumn{1}{c}{854.97} &\multicolumn{1}{c}{1.11} & \multicolumn{1}{c}{3211.49} &\multicolumn{1}{c}{1.08}  \\ 
\toprule
\hfil\multirow{6}{*}{20}\hfill & \multicolumn{1}{c}{0.0} & \multicolumn{1}{c}{161.09} &\multicolumn{1}{c}{76.61} & \multicolumn{1}{c}{108.75} &\multicolumn{1}{c}{79.14} & \multicolumn{1}{c}{0.0} & \multicolumn{1}{c}{182.06} &\multicolumn{1}{c}{45.71} & \multicolumn{1}{c}{214.51} &\multicolumn{1}{c}{51.71}  \\ 
& \multicolumn{1}{c}{1e-05} & \multicolumn{1}{c}{174.62} &\multicolumn{1}{c}{75.98} & \multicolumn{1}{c}{116.3} &\multicolumn{1}{c}{79.98} & \multicolumn{1}{c}{1e-05} & \multicolumn{1}{c}{174.36} &\multicolumn{1}{c}{46.69} & \multicolumn{1}{c}{201.91} &\multicolumn{1}{c}{52.07}  \\ 
& \multicolumn{1}{c}{0.0001} & \multicolumn{1}{c}{142.46} &\multicolumn{1}{c}{76.94} & \multicolumn{1}{c}{141.75} &\multicolumn{1}{c}{79.51} & \multicolumn{1}{c}{0.0001} & \multicolumn{1}{c}{191.54} & \multicolumn{1}{c}{46.53} & \multicolumn{1}{c}{206.77} & \multicolumn{1}{c}{51.78}  \\ 
& \multicolumn{1}{c}{0.001} & \multicolumn{1}{c}{205.24} &\multicolumn{1}{c}{76.11} & \multicolumn{1}{c}{137.72} &\multicolumn{1}{c}{80.24} & \multicolumn{1}{c}{0.001} & \multicolumn{1}{c}{171.26} &\multicolumn{1}{c}{47.29} & \multicolumn{1}{c}{214.15} &\multicolumn{1}{c}{52.35}  \\ 
& \multicolumn{1}{c}{0.01} & \multicolumn{1}{c}{421.03} &\multicolumn{1}{c}{67.73} & \multicolumn{1}{c}{346.64} &\multicolumn{1}{c}{79.93} & \multicolumn{1}{c}{0.01} & \multicolumn{1}{c}{458.96} &\multicolumn{1}{c}{49.53} & \multicolumn{1}{c}{361.27} &\multicolumn{1}{c}{55.18}  \\ 
& \multicolumn{1}{c}{0.1} & \multicolumn{1}{c}{86.73} &\multicolumn{1}{c}{77.95} & \multicolumn{1}{c}{176.45} &\multicolumn{1}{c}{80.85} & \multicolumn{1}{c}{0.1} & \multicolumn{1}{c}{\NA} &\multicolumn{1}{c}{7.0} & \multicolumn{1}{c}{383.14} &\multicolumn{1}{c}{24.72}  \\ 
\toprule
\hfil\multirow{6}{*}{40}\hfill & \multicolumn{1}{c}{0.0} & \multicolumn{1}{c}{152.67} &\multicolumn{1}{c}{76.94} & \multicolumn{1}{c}{106.7} &\multicolumn{1}{c}{81.71} & \multicolumn{1}{c}{0.0} & \multicolumn{1}{c}{169.7} &\multicolumn{1}{c}{44.68} & \multicolumn{1}{c}{214.5} &\multicolumn{1}{c}{51.4}  \\ 
& \multicolumn{1}{c}{1e-05} & \multicolumn{1}{c}{176.08} &\multicolumn{1}{c}{76.71} & \multicolumn{1}{c}{111.17} &\multicolumn{1}{c}{82.5} & \multicolumn{1}{c}{1e-05} & \multicolumn{1}{c}{177.19} &\multicolumn{1}{c}{45.36} & \multicolumn{1}{c}{221.92} &\multicolumn{1}{c}{51.74}  \\ 
& \multicolumn{1}{c}{0.0001} & \multicolumn{1}{c}{162.47} &\multicolumn{1}{c}{76.13} & \multicolumn{1}{c}{\NA} & \multicolumn{1}{c}{\NA} & \multicolumn{1}{c}{0.0001} & \multicolumn{1}{c}{201.82} & \multicolumn{1}{c}{45.89} & \multicolumn{1}{c}{225.43} & \multicolumn{1}{c}{52.91}  \\ 
& \multicolumn{1}{c}{0.001} & \multicolumn{1}{c}{159.88} &\multicolumn{1}{c}{76.18} & \multicolumn{1}{c}{122.83} &\multicolumn{1}{c}{82.48} & \multicolumn{1}{c}{0.001} & \multicolumn{1}{c}{235.74} &\multicolumn{1}{c}{46.37} & \multicolumn{1}{c}{228.98} &\multicolumn{1}{c}{53.58}  \\ 
& \multicolumn{1}{c}{0.01} & \multicolumn{1}{c}{189.76} &\multicolumn{1}{c}{71.75} & \multicolumn{1}{c}{224.72} &\multicolumn{1}{c}{80.93} & \multicolumn{1}{c}{0.01} & \multicolumn{1}{c}{265.38} &\multicolumn{1}{c}{45.76} & \multicolumn{1}{c}{271.12} &\multicolumn{1}{c}{60.22}  \\ 
& \multicolumn{1}{c}{0.1} & \multicolumn{1}{c}{101.32} &\multicolumn{1}{c}{81.28} & \multicolumn{1}{c}{182.46} &\multicolumn{1}{c}{75.73} & \multicolumn{1}{c}{0.1} & \multicolumn{1}{c}{361.16} &\multicolumn{1}{c}{5.4} & \multicolumn{1}{c}{290.84} &\multicolumn{1}{c}{20.5}  \\ 
\bottomrule
\end{tabular}
}
\caption{Unlearning error and testing accuracy for 4 different settings for varying regularization strengths of the $\ell_2$ regularizer and pretraining amount compared to zero regularization.} 
\label{table: l2 reg}
\end{table*}

\begin{table*}[ht!]
\large
\resizebox{\linewidth}{!}{
\begin{tabular}{ccSSSScSSSS}
\toprule
\multirow{2}{*}{\textbf{Pretrain Epochs}} & 
\multirow{1}{*}{\textbf{Regularization}} & 
\multicolumn{2}{c}{\textbf{ResNet-18 CIFAR-10}} & 
\multicolumn{2}{c}{\textbf{VGG-19 CIFAR-10}} & 
\multirow{1}{*}{\textbf{Regularization}} & 
\multicolumn{2}{c}{\textbf{ResNet-18 CIFAR-100}} & 
\multicolumn{2}{c}{\textbf{VGG-19 CIFAR-100}} \\ 
 & \textbf{CIFAR-10} & \textbf{Unlearning Error} & \textbf{Accuracy (\%)} & \textbf{Unlearning Error} &\textbf{Accuracy (\%) } &\textbf{CIFAR-100} & \textbf{Unlearning Error} & \textbf{Accuracy (\%) } & \textbf{Unlearning Error} & \textbf{Accuracy (\%)} \\ 
\midrule 
\hfil\multirow{7}{*}{0}\hfill & \multicolumn{1}{c}{0.0} & \multicolumn{1}{c}{141.35} &\multicolumn{1}{c}{8.82} & \multicolumn{1}{c}{160.95} &\multicolumn{1}{c}{9.03} & \multicolumn{1}{c}{0.0} & \multicolumn{1}{c}{165.65} &\multicolumn{1}{c}{0.78} & \multicolumn{1}{c}{245.66} &\multicolumn{1}{c}{1.09}  \\ 
& \multicolumn{1}{c}{1.0} & \multicolumn{1}{c}{110.32} &\multicolumn{1}{c}{9.66} & \multicolumn{1}{c}{348.85} &\multicolumn{1}{c}{10.51} & \multicolumn{1}{c}{50.0} & \multicolumn{1}{c}{142.59} &\multicolumn{1}{c}{1.01} & \multicolumn{1}{c}{381.88} &\multicolumn{1}{c}{1.38}  \\ 
& \multicolumn{1}{c}{5.0} & \multicolumn{1}{c}{102.71} &\multicolumn{1}{c}{9.82} & \multicolumn{1}{c}{139.47} &\multicolumn{1}{c}{9.98} & \multicolumn{1}{c}{100.0} & \multicolumn{1}{c}{127.31} &\multicolumn{1}{c}{0.99} & \multicolumn{1}{c}{288.04} &\multicolumn{1}{c}{1.0}  \\ 
& \multicolumn{1}{c}{10.0} & \multicolumn{1}{c}{90.66} &\multicolumn{1}{c}{10.06} & \multicolumn{1}{c}{203.44} &\multicolumn{1}{c}{10.15} & \multicolumn{1}{c}{150.0} & \multicolumn{1}{c}{110.48} &\multicolumn{1}{c}{0.95} & \multicolumn{1}{c}{186.17} &\multicolumn{1}{c}{1.0}  \\ 
& \multicolumn{1}{c}{15.0} & \multicolumn{1}{c}{44.85} &\multicolumn{1}{c}{10.61} & \multicolumn{1}{c}{156.59} &\multicolumn{1}{c}{10.07} & \multicolumn{1}{c}{200.0} & \multicolumn{1}{c}{58.59} &\multicolumn{1}{c}{0.74} & \multicolumn{1}{c}{48.08} &\multicolumn{1}{c}{0.85}  \\ 
& \multicolumn{1}{c}{20.0} & \multicolumn{1}{c}{21.63} &\multicolumn{1}{c}{10.32} & \multicolumn{1}{c}{69.34} &\multicolumn{1}{c}{9.5} & \multicolumn{1}{c}{250.0} & \multicolumn{1}{c}{30.81} &\multicolumn{1}{c}{0.85} & \multicolumn{1}{c}{34.56} &\multicolumn{1}{c}{1.21}  \\ 
\toprule
\hfil\multirow{7}{*}{10}\hfill & \multicolumn{1}{c}{0.0} & \multicolumn{1}{c}{132.49} &\multicolumn{1}{c}{77.11} & \multicolumn{1}{c}{141.73} &\multicolumn{1}{c}{77.69} & \multicolumn{1}{c}{0.0} & \multicolumn{1}{c}{251.12} &\multicolumn{1}{c}{44.07} & \multicolumn{1}{c}{258.63} &\multicolumn{1}{c}{48.39}  \\ 
& \multicolumn{1}{c}{1.0} & \multicolumn{1}{c}{101.79} &\multicolumn{1}{c}{74.5} & \multicolumn{1}{c}{123.17} &\multicolumn{1}{c}{76.81} & \multicolumn{1}{c}{50.0} & \multicolumn{1}{c}{176.35} &\multicolumn{1}{c}{43.52} & \multicolumn{1}{c}{233.65} &\multicolumn{1}{c}{50.08}  \\ 
& \multicolumn{1}{c}{5.0} & \multicolumn{1}{c}{98.3} &\multicolumn{1}{c}{73.21} & \multicolumn{1}{c}{113.27} &\multicolumn{1}{c}{77.16} & \multicolumn{1}{c}{100.0} & \multicolumn{1}{c}{177.28} &\multicolumn{1}{c}{43.12} & \multicolumn{1}{c}{212.71} &\multicolumn{1}{c}{47.49}  \\ 
& \multicolumn{1}{c}{10.0} & \multicolumn{1}{c}{90.71} &\multicolumn{1}{c}{69.13} & \multicolumn{1}{c}{107.9} &\multicolumn{1}{c}{74.07} & \multicolumn{1}{c}{150.0} & \multicolumn{1}{c}{171.77} &\multicolumn{1}{c}{38.5} & \multicolumn{1}{c}{216.0} &\multicolumn{1}{c}{46.97}  \\ 
& \multicolumn{1}{c}{15.0} & \multicolumn{1}{c}{86.17} &\multicolumn{1}{c}{68.46} & \multicolumn{1}{c}{78.82} &\multicolumn{1}{c}{65.32} & \multicolumn{1}{c}{200.0} & \multicolumn{1}{c}{124.74} &\multicolumn{1}{c}{26.68} & \multicolumn{1}{c}{195.89} &\multicolumn{1}{c}{36.75}  \\ 
& \multicolumn{1}{c}{20.0} & \multicolumn{1}{c}{11.5} &\multicolumn{1}{c}{34.74} & \multicolumn{1}{c}{27.94} &\multicolumn{1}{c}{31.18} & \multicolumn{1}{c}{250.0} & \multicolumn{1}{c}{34.53} &\multicolumn{1}{c}{13.76} & \multicolumn{1}{c}{59.9} &\multicolumn{1}{c}{8.98}  \\ 
\toprule
\hfil\multirow{7}{*}{20}\hfill & \multicolumn{1}{c}{0.0} & \multicolumn{1}{c}{153.78} &\multicolumn{1}{c}{76.34} & \multicolumn{1}{c}{120.19} &\multicolumn{1}{c}{79.05} & \multicolumn{1}{c}{0.0} & \multicolumn{1}{c}{194.65} &\multicolumn{1}{c}{45.79} & \multicolumn{1}{c}{260.59} &\multicolumn{1}{c}{52.27}  \\ 
& \multicolumn{1}{c}{1.0} & \multicolumn{1}{c}{109.07} &\multicolumn{1}{c}{74.14} & \multicolumn{1}{c}{135.44} &\multicolumn{1}{c}{80.93} & \multicolumn{1}{c}{50.0} & \multicolumn{1}{c}{140.1} &\multicolumn{1}{c}{45.77} & \multicolumn{1}{c}{220.46} &\multicolumn{1}{c}{52.55}  \\ 
& \multicolumn{1}{c}{5.0} & \multicolumn{1}{c}{90.61} &\multicolumn{1}{c}{74.01} & \multicolumn{1}{c}{93.58} &\multicolumn{1}{c}{79.97} & \multicolumn{1}{c}{100.0} & \multicolumn{1}{c}{165.91} &\multicolumn{1}{c}{45.45} & \multicolumn{1}{c}{208.58} &\multicolumn{1}{c}{53.12}  \\ 
& \multicolumn{1}{c}{10.0} & \multicolumn{1}{c}{86.31} &\multicolumn{1}{c}{69.88} & \multicolumn{1}{c}{100.09} &\multicolumn{1}{c}{78.21} & \multicolumn{1}{c}{150.0} & \multicolumn{1}{c}{181.78} &\multicolumn{1}{c}{44.89} & \multicolumn{1}{c}{233.99} &\multicolumn{1}{c}{52.42}  \\ 
& \multicolumn{1}{c}{15.0} & \multicolumn{1}{c}{86.67} &\multicolumn{1}{c}{68.69} & \multicolumn{1}{c}{95.62} &\multicolumn{1}{c}{73.81} & \multicolumn{1}{c}{200.0} & \multicolumn{1}{c}{184.81} &\multicolumn{1}{c}{41.18} & \multicolumn{1}{c}{241.61} &\multicolumn{1}{c}{47.34}  \\ 
& \multicolumn{1}{c}{20.0} & \multicolumn{1}{c}{10.74} &\multicolumn{1}{c}{36.29} & \multicolumn{1}{c}{32.23} &\multicolumn{1}{c}{38.88} & \multicolumn{1}{c}{250.0} & \multicolumn{1}{c}{116.28} &\multicolumn{1}{c}{27.7} & \multicolumn{1}{c}{131.24} &\multicolumn{1}{c}{23.15}  \\ 
\toprule
\hfil\multirow{7}{*}{30}\hfill & \multicolumn{1}{c}{0.0} & \multicolumn{1}{c}{153.26} &\multicolumn{1}{c}{76.48} & \multicolumn{1}{c}{113.37} &\multicolumn{1}{c}{81.5} & \multicolumn{1}{c}{0.0} & \multicolumn{1}{c}{210.35} &\multicolumn{1}{c}{45.68} & \multicolumn{1}{c}{250.87} &\multicolumn{1}{c}{52.17}  \\ 
& \multicolumn{1}{c}{1.0} & \multicolumn{1}{c}{112.01} &\multicolumn{1}{c}{73.7} & \multicolumn{1}{c}{130.25} &\multicolumn{1}{c}{81.18} & \multicolumn{1}{c}{50.0} & \multicolumn{1}{c}{151.56} &\multicolumn{1}{c}{45.02} & \multicolumn{1}{c}{194.86} &\multicolumn{1}{c}{52.69}  \\ 
& \multicolumn{1}{c}{5.0} & \multicolumn{1}{c}{94.44} &\multicolumn{1}{c}{72.48} & \multicolumn{1}{c}{108.84} &\multicolumn{1}{c}{80.24} & \multicolumn{1}{c}{100.0} & \multicolumn{1}{c}{178.74} &\multicolumn{1}{c}{45.57} & \multicolumn{1}{c}{229.36} &\multicolumn{1}{c}{52.88}  \\ 
& \multicolumn{1}{c}{10.0} & \multicolumn{1}{c}{83.53} &\multicolumn{1}{c}{70.19} & \multicolumn{1}{c}{116.96} &\multicolumn{1}{c}{78.76} & \multicolumn{1}{c}{150.0} & \multicolumn{1}{c}{205.75} &\multicolumn{1}{c}{42.71} & \multicolumn{1}{c}{217.5} &\multicolumn{1}{c}{52.48}  \\ 
& \multicolumn{1}{c}{15.0} & \multicolumn{1}{c}{88.96} &\multicolumn{1}{c}{69.17} & \multicolumn{1}{c}{92.93} &\multicolumn{1}{c}{77.13} & \multicolumn{1}{c}{200.0} & \multicolumn{1}{c}{203.18} &\multicolumn{1}{c}{43.1} & \multicolumn{1}{c}{256.51} &\multicolumn{1}{c}{51.81}  \\ 
& \multicolumn{1}{c}{20.0} & \multicolumn{1}{c}{13.73} &\multicolumn{1}{c}{36.98} & \multicolumn{1}{c}{49.03} &\multicolumn{1}{c}{48.69} & \multicolumn{1}{c}{250.0} & \multicolumn{1}{c}{127.97} &\multicolumn{1}{c}{28.52} & \multicolumn{1}{c}{200.4} &\multicolumn{1}{c}{37.02}  \\ 
\toprule
\hfil\multirow{7}{*}{40}\hfill & \multicolumn{1}{c}{0.0} & \multicolumn{1}{c}{175.32} &\multicolumn{1}{c}{75.59} & \multicolumn{1}{c}{120.92} &\multicolumn{1}{c}{82.07} & \multicolumn{1}{c}{0.0} & \multicolumn{1}{c}{209.52} &\multicolumn{1}{c}{45.11} & \multicolumn{1}{c}{219.14} &\multicolumn{1}{c}{52.32}  \\ 
& \multicolumn{1}{c}{1.0} & \multicolumn{1}{c}{97.11} &\multicolumn{1}{c}{74.86} & \multicolumn{1}{c}{109.44} &\multicolumn{1}{c}{80.89} & \multicolumn{1}{c}{50.0} & \multicolumn{1}{c}{176.31} &\multicolumn{1}{c}{45.45} & \multicolumn{1}{c}{250.64} &\multicolumn{1}{c}{52.36}  \\ 
& \multicolumn{1}{c}{5.0} & \multicolumn{1}{c}{93.74} &\multicolumn{1}{c}{73.58} & \multicolumn{1}{c}{108.54} &\multicolumn{1}{c}{80.21} & \multicolumn{1}{c}{100.0} & \multicolumn{1}{c}{201.74} &\multicolumn{1}{c}{45.01} & \multicolumn{1}{c}{251.3} &\multicolumn{1}{c}{53.01}  \\ 
& \multicolumn{1}{c}{10.0} & \multicolumn{1}{c}{83.11} &\multicolumn{1}{c}{70.33} & \multicolumn{1}{c}{92.38} &\multicolumn{1}{c}{79.31} & \multicolumn{1}{c}{150.0} & \multicolumn{1}{c}{211.01} &\multicolumn{1}{c}{42.12} & \multicolumn{1}{c}{233.39} &\multicolumn{1}{c}{52.49}  \\ 
& \multicolumn{1}{c}{15.0} & \multicolumn{1}{c}{79.86} &\multicolumn{1}{c}{68.7} & \multicolumn{1}{c}{96.14} &\multicolumn{1}{c}{77.64} & \multicolumn{1}{c}{200.0} & \multicolumn{1}{c}{218.61} &\multicolumn{1}{c}{41.58} & \multicolumn{1}{c}{233.58} &\multicolumn{1}{c}{50.67}  \\ 
& \multicolumn{1}{c}{20.0} & \multicolumn{1}{c}{8.24} &\multicolumn{1}{c}{41.02} & \multicolumn{1}{c}{58.35} &\multicolumn{1}{c}{44.14} & \multicolumn{1}{c}{250.0} & \multicolumn{1}{c}{198.57} &\multicolumn{1}{c}{41.05} & \multicolumn{1}{c}{253.88} &\multicolumn{1}{c}{45.69}  \\ 
\toprule
\hfil\multirow{7}{*}{50}\hfill & \multicolumn{1}{c}{0.0} & \multicolumn{1}{c}{184.99} &\multicolumn{1}{c}{76.89} & \multicolumn{1}{c}{118.56} &\multicolumn{1}{c}{81.39} & \multicolumn{1}{c}{0.0} & \multicolumn{1}{c}{235.23} &\multicolumn{1}{c}{45.51} & \multicolumn{1}{c}{221.19} &\multicolumn{1}{c}{52.47}  \\ 
& \multicolumn{1}{c}{1.0} & \multicolumn{1}{c}{100.39} &\multicolumn{1}{c}{74.51} & \multicolumn{1}{c}{121.06} &\multicolumn{1}{c}{81.37} & \multicolumn{1}{c}{50.0} & \multicolumn{1}{c}{168.65} &\multicolumn{1}{c}{44.91} & \multicolumn{1}{c}{226.56} &\multicolumn{1}{c}{51.95}  \\ 
& \multicolumn{1}{c}{5.0} & \multicolumn{1}{c}{88.43} &\multicolumn{1}{c}{73.57} & \multicolumn{1}{c}{114.11} &\multicolumn{1}{c}{80.31} & \multicolumn{1}{c}{100.0} & \multicolumn{1}{c}{206.35} &\multicolumn{1}{c}{44.57} & \multicolumn{1}{c}{224.88} &\multicolumn{1}{c}{52.67}  \\ 
& \multicolumn{1}{c}{10.0} & \multicolumn{1}{c}{80.72} &\multicolumn{1}{c}{72.14} & \multicolumn{1}{c}{113.41} &\multicolumn{1}{c}{79.28} & \multicolumn{1}{c}{150.0} & \multicolumn{1}{c}{229.16} &\multicolumn{1}{c}{43.73} & \multicolumn{1}{c}{230.4} &\multicolumn{1}{c}{53.57}  \\ 
& \multicolumn{1}{c}{15.0} & \multicolumn{1}{c}{76.42} &\multicolumn{1}{c}{71.39} & \multicolumn{1}{c}{87.12} &\multicolumn{1}{c}{77.17} & \multicolumn{1}{c}{200.0} & \multicolumn{1}{c}{234.73} &\multicolumn{1}{c}{43.04} & \multicolumn{1}{c}{255.56} &\multicolumn{1}{c}{51.29}  \\ 
& \multicolumn{1}{c}{20.0} & \multicolumn{1}{c}{17.43} &\multicolumn{1}{c}{37.31} & \multicolumn{1}{c}{77.01} &\multicolumn{1}{c}{68.4} & \multicolumn{1}{c}{250.0} & \multicolumn{1}{c}{228.22} &\multicolumn{1}{c}{40.7} & \multicolumn{1}{c}{254.52} &\multicolumn{1}{c}{47.52}  \\ 
\bottomrule
\end{tabular}
}
\caption{Unlearning error and testing accuracy for 4 different settings for varying regularization strengths of the Standard Deviation regularizer and pretraining amount. \textcolor{black}{Note $t=1$ epoch here.}} 
\label{tb:SD_loss_unl_err}
\end{table*}

\end{document}